\documentclass[twoside,11pt]{article}

\usepackage[abbrvbib, preprint]{jmlr2e}

\usepackage{lastpage}
\jmlrheading{23}{2022}{1-\pageref{LastPage}}{1/21; Revised 5/22}{9/22}{21-0000}{Author One and Author Two}

\ShortHeadings{A Markov Random Field model for Hypergraph-based Machine Learning}{Bohan Tang, Keyue Jiang, Laura Toni, Siheng Chen and Xiaowen Dong}

\usepackage{amsmath,amsfonts,bm}

\usepackage{blindtext}

\usepackage{xcolor}

\newcommand{\R}{\ensuremath{\mathbb{R}}}

\def\y{\mathbf{y}}
\def\hH{\mathbf{H}}

\def\E{\mathcal{E}}
\def\hhH{\mathcal{H}}
\def\G{\mathcal{G}}

\def\z{\mathbf{z}}
\def\Z{\mathbf{Z}}

\def\y{\mathbf{y}}

\def\N{\mathcal{N}}

\def\K{\mathcal{K}}

\DeclareMathOperator{\h}{h}

\DeclareMathOperator{\mmin}{min}

\def\eqref#1{equation~\ref{#1}}

\def\1{\bm{1}}

\def\vh{{\bm{h}}}

\def\vs{{\bm{s}}}

\def\vw{{\bm{w}}}
\def\vx{{\bm{x}}}

\def\vz{{\bm{z}}}

\def\mH{{\bm{H}}}

\def\mL{{\bm{L}}}

\def\mW{{\bm{W}}}
\def\mX{{\bm{X}}}
\def\mY{{\bm{Y}}}
\def\mZ{{\bm{Z}}}

\DeclareMathAlphabet{\mathsfit}{\encodingdefault}{\sfdefault}{m}{sl}
\SetMathAlphabet{\mathsfit}{bold}{\encodingdefault}{\sfdefault}{bx}{n}

\def\gE{{\mathcal{E}}}

\def\gG{{\mathcal{G}}}
\def\gH{{\mathcal{H}}}

\def\gJ{{\mathcal{J}}}

\def\gV{{\mathcal{V}}}

\def\sR{{\mathbb{R}}}

\DeclareMathOperator*{\argmax}{arg\,max}
\DeclareMathOperator*{\argmin}{arg\,min}

\usepackage{amsmath,amsfonts}

\usepackage{textcomp}
\usepackage{stfloats}
\usepackage{url}
\usepackage{verbatim}
\usepackage{hyperref}
\usepackage{cleveref}

\usepackage{graphicx}
\usepackage{subcaption} 
\usepackage{floatflt} 
\usepackage{paralist} 
\usepackage[ruled]{algorithm2e}
\usepackage{algorithmic,algorithm2e}
\usepackage{amssymb}
\usepackage{xcolor}
\usepackage{booktabs}
\usepackage{multirow}

\def\Z{\mathbf{Z}}

\def\z{\mathbf{z}}

\def\y{\mathbf{y}}

\def\V{\mathcal{V}}
\def\E{\mathcal{E}}
\def\K{\mathcal{K}}
\def\m{\mathcal{M}}
\def\gH{\mathcal{H}}

\def\G{\mathcal{G}}

\def\N{\mathcal{N}}
\def\cals{\mathcal{S}}
\def\ttt{\mathbf{\Theta}}

\DeclareMathOperator{\diag}{diag}
\DeclareMathOperator{\st}{s.t.}

\begin{document}

\title{A Markov Random Field model for Hypergraph-based Machine Learning}

\author{\name Bohan Tang\thanks{Equal Contribution.} \email bohan.tang@eng.ox.ac.uk \\
       \addr Department of Engineering Science\\
       University of Oxford\\
       Oxford, UK
       \AND
       \name Keyue Jiang$^*$ \email keyue.jiang.18@ucl.ac.uk \\
       \addr Department of Electronic and Electrical Engineering\\
       University College London\\
       London, UK
       \AND
       \name Laura Toni \email l.toni@ucl.ac.uk \\
       \addr Department of Electronic and Electrical Engineering\\
       University College London\\
       London, UK
       \AND
       \name Siheng Chen \email sihengc@sjtu.edu.cn \\
       \addr Shanghai Jiao Tong University and Shanghai AI laboratory\\
    Shanghai, China
       \AND
       \name Xiaowen Dong \email xdong@robots.ox.ac.uk \\
       \addr Department of Engineering Science\\
       University of Oxford\\
       Oxford, UK
       }
\editor{My editor}

\maketitle

\begin{abstract}
Understanding the data-generating process is essential for building machine learning models that generalise well while ensuring robustness and interpretability. This paper addresses the fundamental challenge of modelling the data generation processes on hypergraphs and explores how such models can inform the design of machine learning algorithms for hypergraph data. The key to our approach is the development of a hypergraph Markov random field that models the joint distribution of the node features and hyperedge features in a hypergraph through a multivariate Gaussian distribution whose covariance matrix is uniquely determined by the hypergraph structure. The proposed data-generating process provides a valuable inductive bias for various hypergraph machine learning tasks, thus enhancing the algorithm design. In this paper, we focus on two representative downstream tasks: structure inference and node classification. Accordingly, we introduce two novel frameworks: 1) an original hypergraph structure inference framework named HGSI, and 2) a novel learning framework entitled Hypergraph-MLP for node classification on hypergraphs. Empirical evaluation of the proposed frameworks demonstrates that: 1) HGSI outperforms existing hypergraph structure inference methods on both synthetic and real-world data; and 2) Hypergraph-MLP outperforms baselines in six hypergraph node classification benchmarks, at the same time promoting runtime efficiency and robustness against structural perturbations during inference.

\end{abstract} 


\section{Introduction}
\label{sec:intro}


Higher-order interactions that involve more than two entities widely exist in various domains, such as co-authorships in scientific collaboration~\cite{han2009understanding}, group contact and spreading phenomena in epidemiology~\cite{jhun2021effective}, and multi-tissue gene expression in biology~\cite{vinas2023hypergraph}. Hypergraphs, characterised by nodes representing entities and hyperedges denoting the higher-order interactions among these entities, serve as a natural tool for modelling real-world data with complex higher-order interactions~\cite{bick2023higher}. The analysis of data collected in hypergraphs has recently garnered growing interest in the machine learning community~\cite{antelmi2023survey,9264674}, leading to tasks such as node classification~\cite{hypergraphmlp} and link prediction~\cite{ali2021improving}.

Like traditional machine learning (ML) tasks, the effectiveness of a model for hypergraph ML depends on the compatibility between the predictive model and the underlying data-generating process (DGP)~\cite{hastie01statisticallearning, DBLP:journals/corr/abs-2011-15091}. 
However, current research on hypergraph machine learning~\cite{DBLP:conf/iclr/KaurK023, 10.1145/3637528.3671457, yang2025recentadvanceshypergraphneural} primarily focuses on designing predictive models, such as hypergraph neural networks~\cite{chien2022you,feng2019hypergraph, 10.1145/3719002}, without investigating the inductive biases induced by the DGP. In parallel to these studies, we seek to advance hypergraph ML from a different perspective - by explicitly modelling the generation process for hypergraph data. To this end, we propose a DGP that incorporates our belief on the hypergraph data formulation and obtain hypergraph Markov random fields (HMRF), an energy-based model borrowing ideas from Gaussian Markov random fields~\cite{zhu2003semi}. Through injecting inductive biases into the solution space of downstream tasks, HMRF evokes new architecture and motivates effective model designs for multiple hypergraph ML tasks, such as structure inference and node classification.

\begin{figure}[t] 
\centering
\includegraphics[width=.8\textwidth]{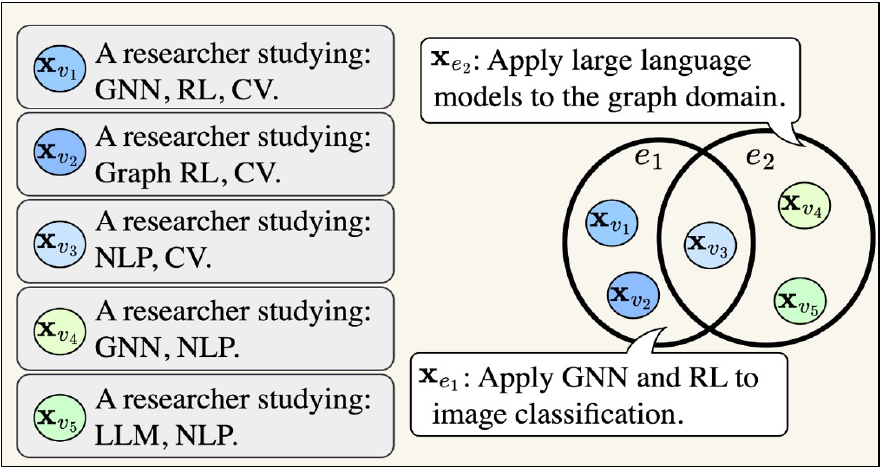}
\caption{A hypergraph of scientific collaborations, where nodes represent authors, node features are the author's research interests, hyperedges represent papers and hyperedge features are the paper's research topics. In this hypergraph, co-authors (nodes) tend to share similar research interests (node features) aligned with their paper's topic (hyperedge features).}
\label{fig:smoothness_prior}
\end{figure}

The proposed HMRF model is rooted in the assumption that the features of nodes in a hyperedge are highly correlated via the features of the hyperedge connecting them (See~\Cref{fig:smoothness_prior} for an example of a hypergraph satisfying this assumption). Under such an assumption, the DGP of a hypergraph is modelled as an energy-based model whose negative exponent (energy function) is uniquely determined by the given hypergraph structure and the joint node-hyperedge features. This induces a multivariate Gaussian distribution when the hypergraph structure is given. Since the hyperedge features are typically unavailable in real-world applications~\cite{chien2022you, feng2019hypergraph},   we approximate the energy function with a computable estimate that does not require explicit hyperedge features, thereby enhancing the applicability of the proposed HMRF. Specifically, the proposed estimate is defined as the sum of the maximum $\ell_2$ distances between  any pair of node features within each hyperedge of a hypergraph. To demonstrate the practical value of our proposed HMRF, we apply it to two key downstream tasks, \textit{hypergraph structure inference} and \textit{hypergraph node classification}. In each task, the energy estimate can be seamlessly incorporated into the optimisation objective as inductive biases to promote better solutions.

The hypergraph structure inference (HGSI) task aims to infer the underlying hypergraph structures from the observed node features,  when the structure is not directly observable. A concrete example of HGSI involves extracting co-authorship relationships by analysing the profile of each individual, such as research interest, affiliation, etc. For this task, we introduce a novel framework named Hypergraph Structure Inference. Utilising observed node features as input, this framework infers probabilities of potential hyperedges by solving a maximum a-posteriori problem evaluated by the HMRF. Compared with previous hypergraph structure inference methods, the key novelty of our proposed approach lies in its capability to directly learn probabilities for potential hyperedges, removing the need for supervision on any pre-defined hypergraphs or hyperedges. 

The second task considered in this work is hypergraph node classification, aimed at  classifying nodes on a pre-defined hypergraph structure. Examples include classifying research fields of academic authors in a co-authorship hypergraph based on published papers, and predicting political alignment in an online discussion forum based on multi-user thread participation. For this task, we design a novel learning framework called Hypergraph-MLP. This framework consists of two primary components: a multi-layer perceptron (MLP) and a loss function rooted in the HMRF which regularises the embedding optimisation through the prior belief adhering to the data-generating process. Training with this loss function enables the MLP to leverage structural information efficiently without message passing. This leads to lower inference latency for Hypergraph-MLP compared to traditional message-passing hypergraph neural networks. Moreover, removing the reliance on the hypergraph structure at inference time makes Hypergraph-MLP robust to structural perturbations.

The main contributions of this work are as follows:
\begin{itemize}
    \item We generalise the classical Markov random fields to model the data-generating process on hypergraphs, yielding a model named HMRF. We extend the applicability of the HMRF to the scenarios when hyperedge features are unobservable through a node-feature based energy estimate.

    \item We develop a novel unsupervised approach for hypergraph structure inference using the proposed HMRF. The unsupervised nature of our method enables it to be applied to scenarios where pre-defined hypergraphs are unavailable.

    \item We design Hypergraph-MLP, a novel learning framework for node classification in pre-defined hypergraph structures based on the proposed HMRF. More generally, it provides a new paradigm for designing neural networks to process hypergraph-structured data without the need for message-passing, thus alleviating inference-time computation and increasing robustness against structural perturbations. 

    \item We conduct extensive experiments to show that: 1) On both synthetic and real-world datasets, HGSI significantly outperforms existing hypergraph structure inference methods; and 2) On six hypergraph node classification benchmarks, compared to existing hypergraph neural networks, Hypergraph-MLP achieves competitive accuracy, fastest inference, and better robustness against structural perturbations at inference.
\end{itemize}

A preliminary version of the Hypergraph-MLP method was proposed and empirically tested in~\cite{hypergraphmlp}. The present version significantly extends the previous work by proposing a novel DGP and HMRF with theoretical grounding, applying it to develop novel frameworks for two different learning tasks, and providing comprehensive empirical validation. 

The rest of this paper is structured as follows. We first review related work in~\Cref{sec:Related_Work}. Then, in~\Cref{sec:preliminary}, we introduce the necessary notations and formalise the hypergraph machine learning tasks from a probabilistic perspective. In~\Cref{sec:hypergraph_smoothness_prior}, we propose a novel hypergraph Markov random field and leverage this model to construct structure likelihood in hypergraph structure inference task and embedding prior in hypergraph node classification tasks. We also develop an energy estimate to tackle the problem when the hyperedge features are unobservable. In~\Cref{sec:hsi}, we elaborate the framework for inferring the structure of an unknown hypergraph. In~\Cref{sec:hnc}, we use the HMRF to design a novel framework for node classification on pre-defined hypergraphs. In~\Cref{sec:Experiments}, we conduct experiments on hypergraph structure inference and hypergraph node classification. Finally, we conclude the paper in~\Cref{sec:Conclusion}. 

\section{Related Work} 
\label{sec:Related_Work}
\subsection{Data-generating Processes for Graph-structured Data}  
Data-generating processes are often introduced for understanding the underlying mechanism that produces the observed data. When modelling the data-generating process for graphs, it is commonly assumed that the features or signals on the graphs admit certain regularity or smoothness according to the graph structure~\cite{DBLP:journals/tsp/DongTFV16, DBLP:conf/aistats/Kalofolias16, DBLP:journals/tsipn/PuCDS21, deshpande2018contextual, jiang2025heterogeneous}. Common choices of the DGPs for graph-structured data are Ising models~\cite{Ising1925BeitragZT}, Gaussian graphical models~\cite{yuan2007model, DBLP:journals/simods/JiaB22}, and pair-wise exponential Markov random fields (PE-MRF)~\cite{DBLP:conf/aistats/ParkHBL17}, where the features are assumed to be emitted from a multivariate Gaussian distribution whose covariance matrix is uniquely determined by the graph structure. Understanding the data-generating process is valuable as it can inspire the design of predictive models~\cite{doi:10.1137/21M1395351} and generative models~\cite{jiang2025bureswassersteinflowmatchinggraph} for graph machine learning, or enhance the interpretability of neural network architecture design~\cite{DBLP:conf/nips/WeiYJB022}. 
To the best of our knowledge, our work is the first investigating the data-generating processes of data collected in hypergraphs and exploring how such prior knowledge can motivate the design of hypergraph machine learning algorithms.

\subsection{Hypergraph Structure Inference} 
Existing hypergraph structure inference approaches can be divided into two categories: rule-based and supervised-learning-based approaches. The rule-based approach either assumes that nodes in a hyperedge have similar features, or that a hyperedge can be decomposed as a set of pairwise edges. The methods built upon the first assumption usually construct a hypergraph by setting hyperedges as node clusters determined by the $k$-means algorithm~\cite{gao2020hypergraph,xu2022GroupNet,xu2022dynamic}. Approaches under the second assumption typically create a hypergraph with specific connected components, e.g., cliques or communities~\cite{Young2021HypergraphRF}. These rule-based methods can produce hypergraph structures with certain desirable properties, but they fail to capture the data-generating process behind the hypergraph structure, which in turn limits their accuracy and robustness at inference time. On the other hand, several recent work~\cite{serviansky2020set2graph,zhang2022pruning} use ground-truth hypergraph structures associated with node features to train a neural network that learns a mapping between node features and ground-truth structures. After the mapping is learned, the neural network can estimate the probabilities of hyperedges and form a hypergraph structure accordingly. However, these supervised-learning-based approaches often require a significant amount of ground-truth hyperedges and associated node features for reliable training, hence limiting their application in scenarios where labelled data are scarce. We address the limitations of both categories in this paper. Leveraging the proposed HMRF, we design an unsupervised hypergraph structure inference method that can estimate the probability for each potential hyperedge without supervision on labelled data.

\subsection{Hypergraph Node Classification} 
There exists a body of work for node classification on pre-defined hypergraphs by utilizing node features and the hypergraph incidence matrix as inputs~\cite{antelmi2023survey}. Most of them~\cite{NEURIPS2019_1efa39bc,feng2019hypergraph,bai2021hypergraph,ijcai21-UniGNN,chien2022you} extend the message-passing paradigm in graph machine learning~\cite{zhou2020graph} to develop hypergraph neural networks. This adaptation employs a two-stage message-passing paradigm: node features are first aggregated to hyperedges to update hyperedge embeddings, which are then aggregated back to nodes to update node embeddings. Despite the promising performance of these message-passing-based models, they have several limitations including high inference latency~\cite{antelmi2023survey,zhang2022graphless}, and sensitivity to structural perturbations~\cite{sun2022adversarial, hu2023hyperattack}. Leveraging our HMRF, we design a framework named Hypergraph-MLP for hypergraph node classification without message passing. Removing the reliance on the hypergraph structure at inference time makes Hypergraph-MLP not only have lower inference latency but also more robust against structural perturbations.

\section{Preliminaries}
\label{sec:preliminary}
We introduce the notations for hypergraphs, and the features of nodes and hyperedges.

\subsection{Notations}

\textbf{Hypergraphs.} A hypergraph can be represented as $\hhH = \{\V, \E, \hH\}$, where $\V = \{v_1, v_2, \cdots, v_n\}$ is the node set with $|\V|=n$, $\E=\{e_1, e_2, \cdots, e_m\}$ is the hyperedge set with $|\E| = m$, and $\hH=[w_1\h_{1}, w_2\h_{2}, \cdots, w_{m}\h_{m}]\in\R^{n\times m}$ is an incidence matrix embedding a hypergraph structure, in which $w_j\in[0,1]$ is the weight of the $j$-th hyperedge and $\h_{j}\in\{0,1\}^{n}$ represents the $j$-th hyperedge: $\hH_{ij}\!=\!1$ indicates that hyperedge $j$ contains node $i$ and $\hH_{ij}\!=\!0$ otherwise. Note that if the hypergraph structure is pre-defined (i.e., all hyperedges are observed), $\h_{j}$ are fixed with all weights set to $1$; if the structure is not observed, $\h_{j}$ would include all possible hyperedges and the task would be to infer the weight (or presence) of each hyperedge. The size of a hyperedge is the number of nodes contained in a hyperedge. 

\textbf{Node and hyperedge features.} For a hypergraph $\gH = \{\gV, \gE, \mH\}$, let $\mX_\gV = [\vx_{v_1}, \vx_{v_2}, \cdots, \vx_{v_n}]^{T} \in \sR^{n\times d}$ denote the node features and $\mX_\gE = [\vx_{e_1}, \vx_{e_2}, \cdots, \vx_{e_m}]^{T} \in \sR^{m\times d}$ denote the hyperedge features, which are two matrices that contain $d$-dimensional features. Notably, when $d=1$, these feature matrices reduce to vectors (i.e., scalar features). To keep the notation concise, we focus our discussion on the case with $d\!=\!1$. However, our results are directly applicable to cases with $d\!>\!1$.

\begin{figure}[t] 
\centering
\includegraphics[width=.5\textwidth]{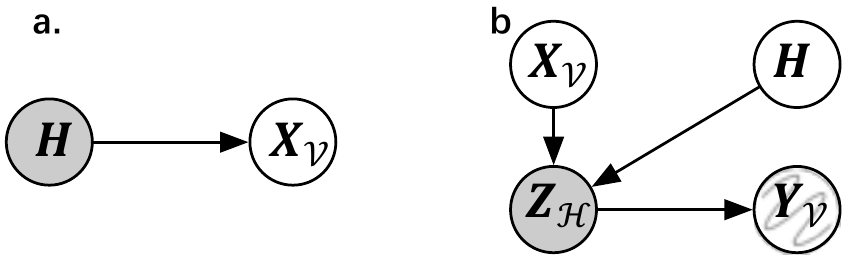}
\caption{(a) The graphical models for hypergraph structure inference, where the task is to infer $\mH$ from $\mX_{\gV}$. (b) The graphical models for hypergraph node classification, where the task is to predict the observed $\mY_{u} \in \mY_{\gV}$. The shaded variables are unobserved and the curve-crossed variables are partially observed.}
\label{fig:graphical_model}
\vspace{-3mm}
\end{figure}

\subsection{Hypergraph Machine Learning Tasks through the lens of Data-generating Processes.} 

The design of the hypergraph ML algorithm is highly related to the underlying data-generating process. We will illustrate the relationship through two tasks: \textit{hypergraph structure inference} and \textit{hypergraph node classification}. This section introduces the general formalisation and the detailed solution is left in~\Cref{sec:hsi} and~\Cref{sec:hnc}.

\textbf{Hypergraph Structure Inference.} The HGSI task aims to take observed node features $\mX_\gV \in \sR^{n\times d}$ as inputs to infer, in an unsupervised fashion, a ground-truth binary incidence matrix $\mH\in\{0,1\}^{n\times m}$ for the hypergraph structure associated with $\mX_\V$. Specifically, HGSI can be formalized as a maximum a-posteriori (MAP) estimation problem:
\begin{equation}
\label{eq: hgsi_formulation}
\begin{aligned}
    \mH^* &= \argmax_\mH P(\mH \mid \mX_\gV) \\
    &=\argmax_\mH \underbrace{\log P(\mX_\gV \mid \mH)}_{\text{log-likelihood}} + \underbrace{\log P(\mH)}_{\text{log-prior}}
\end{aligned}
\end{equation}
where the knowledge of the data-generating process would inform the analytical form of the likelihood function of the structure, i.e., $P(\mX_\gV \mid \mH)$. In~\Cref{sec:hsi} we will further introduce how to turn the MAP problem of Eq.~(\ref{eq: hgsi_formulation}) into a meaningful optimisation problem.

\textbf{Hypergraph Node Classification.} Let $\mX_{\V} \in \sR^{n\times d}$ denote the observed node features, $\V_{l}$ the set of labelled nodes with ground truth labels $\mY_{l}=\{\y_v\}_{v\in\V_{l}}$, where $\y_{v_i}\in\{0,1\}^{c}$ is a one-hot encoding of the node label (from $c$ classes), and $\V_{u}=\V\backslash\V_{l}$ be a set of unlabeled nodes. The hypergraph node classification task requires a model to classify nodes within $\V_{u}$ based on $\mX_\V$, known labels $\mY_{l}$, and the given hypergraph structure $\mH$. The task has two steps: 1) MAP inference of model parameters $\Theta$, where the model can be any neural network:
\begin{equation}
\label{eq: training}
\begin{aligned}
\Theta^* &= \argmax_{\Theta} P\left(  \Theta\mid \mY_{l}, \mX_\gV, \mH \right) \\
&= \argmax_{\Theta} \underbrace{\log P(\mY_{l}\mid \Theta, \mX_\gV, \mH)}_{\text{log-likelihood}}+ \underbrace{\log P\left(\Theta \mid \mX_\gV, \mH \right)}_{\text{log-prior}}
\end{aligned}
\end{equation}
and 2) prediction based on the model with optimised parameters\footnote{To distinguish, we use starred representation $\Theta^*$ for the optimal model parameter and hatted $\hat{\mY}_{un}$ for the inferred labels, i.e., the output of the model.}:
\begin{equation}
\hat{\mY}_{u} = \argmax_{\mY_{u}} P\left(\mY_{u} \mid \mX_\gV,\mH , \Theta^*\right).
\end{equation}
Many existing hypergraph neural networks~\cite{chien2022you,feng2019hypergraph,ijcai21-UniGNN} are designed by a two-step message passing paradigm, i.e., the node embeddings ($\mZ_{\V}$) are sent to its corresponding hyperedges to learn the hyperedge latent embeddings ($\mZ_{\E}$), and then the learned hyperedge embeddings are passed back to its nodes to refine the node embeddings, where the initial node embeddings and the hypergraph structure taken from the datasets. Inspired by the two-step message passing paradigm, we consider the model parameters to be latent node embeddings denoted as $\mZ_{\gH} = [\mZ_{\V}^{T}, \mZ_{\E}^{T}]^{T}$ and consider the graphical model as in~\Cref{fig:graphical_model}b, the model prediction can be decomposed into (we omitted $\Theta$ for simplicity) if assuming $\mZ_\gH$ following a Dirac distribution with point mass.
\begin{equation}
\label{eq: likelihood}
\begin{aligned}
P(\mY_{l}\mid \mX_\gV, \mH) &=  \int_{\mZ_{\gH}} P(\mY_{l}\mid \mZ_\gH)P(\mZ_\gH \mid \mX_\gV, \mH) \\
&= P(\mY_{l}\mid \mZ_\gH)*P(\mZ_\gH \mid \mX_\gV) * P(\mZ_\gH \mid \mH).
\end{aligned}
\end{equation}
We assume 1) $\mX_\gV \perp\mH \mid \mZ_\gH$, and
2) $\mX_\gV$ and $\mH$ provide independent evidence about $\mZ_\gH$. 

Considering the optimal point preservation of log transformation, the training objective in~\Cref{eq: training} becomes,
\begin{equation}
\label{eq:node_class_obj}
\underbrace{\log P(\mY_{l}\mid \mZ_\gH)}_{\text{log-likelihood}}+ \underbrace{\log P\left(\mZ_\gH \mid \mH\right)}_{\text{prior from structure}} +\underbrace{\log P \left( \mZ_\gH \mid \mX_\gV \right)}_{\text{prior from features}}.
\end{equation}


\section{Data-generating processes for hypergraphs}
\label{sec:hypergraph_smoothness_prior}
In this section, we first propose a hypergraph Markov random field for modelling the generation process of hypergraph data and, more specifically, the term $P(\mX_\gV \mid \mH)$ in the structure learning task and $P(\mZ_\gV \mid \mH)$ in the node classification tasks. In our proposed model, despite the difference in interpretations as pointed out above, these two terms share the same analytical form that is extended from the graph Markov random field proposed in~\cite{doi:10.1137/21M1395351}. Based on this, we further develop an energy estimate for the DGP that addresses the challenge of missing hyperedge features. 

\begin{figure}[t]
\begin{center}
\includegraphics[width=.8\columnwidth]{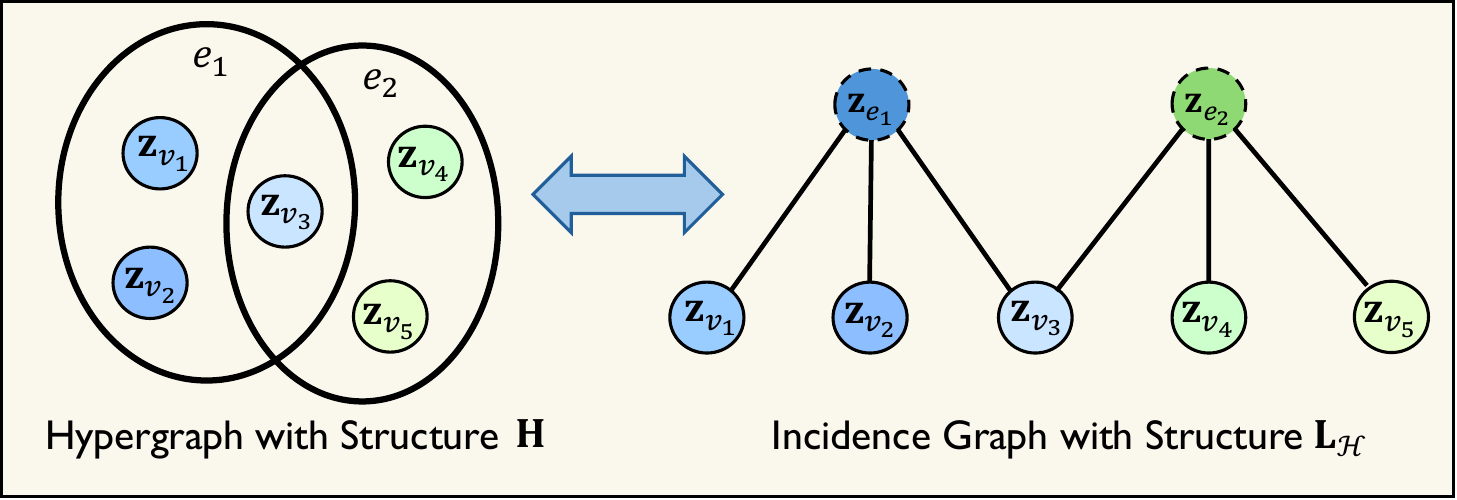}
 \caption[Short caption for the list of figures]{$\mX_\gV = [\vx_{v_1}, \vx_{v_2}, \cdots, \vx_{v_n}]^{T} \in \sR^{n\times d}$ and  $\mX_{\E} = [\vx_{e_1}, \vx_{e_2}, \cdots, \vx_{e_{m}}]^{T} \in \sR^{m\times d}$ are node and hyperedge features respectively. For data on a hypergraph $\mH$, the relationship between $\mX_\V$ and $\mX_\E$ can be modelled by an incidence graph corresponding to $\mH$ with a graph Laplacian $\mL_{\gH}$.}
\label{fig:pf_pipeline}
\end{center}
\end{figure}
\subsection{Hypergraph Markov Random Field} 
\label{sec:gsp}
We start by introducing the fundamental assumption of our hypergraph Markov random field. Building upon this assumption, we represent the relationship between nodes and hyperedges in the feature space through an incidence graph. Finally, leveraging this incidence graph, we present the probabilistic formulation of our hypergraph Markov random field\footnote{The model is ``Markovian'' in the sense that the nodes and hyperedges are conditionally independent of the other nodes and hyperedges given their directly connected components in a node-edge connection.}. This formulation characterises the distribution of node and hyperedge features within a hypergraph as a multivariate Gaussian distribution.

\textbf{Assumption.} To characterise the criteria by which nodes with certain features can be connected via a hyperedge, we assume that: \textit{the features of nodes in a hyperedge are highly correlated via the features of the hyperedge connecting them.} One real-world example satisfying this assumption is a hypergraph of scientific collaboration where nodes represent authors and features the author's research interests, and hyperedges represent papers and features the paper's research topic; See~\Cref{fig:smoothness_prior} for an illustrative example.

\textbf{Incidence graph.} Our assumption indicates that, in the feature space, node embeddings are highly correlated to their corresponding hyperedge embeddings. In this case, we capture the relation between nodes and hyperedges by using an incidence graph~\cite{godsil2001algebraic,bahmanian2015connection,ouvrard2020hypergraphs} corresponding to the hypergraph structure $\mH$. 

Specifically, for the hypergraph $\gH=\{\V,\E,\mH\}$, we construct a unique incidence graph that is a bipartite graph $\G=\{\V\bigcup\V', \E_{\G}, \mL_{\gH}\}$, where $v_i\in\V$ is a node in $\gH$, $v_{e_{j}}\in\V'$ corresponds to a hyperedge $e_j$ in $\gH$, and there is an edge between $v_i$ and $v_{e_{j}}$ if and only if $e_{j}$ contains $v_i$ in $\gH$. The structure of $\gG$ can be represented as a graph Laplacian matrix:
\begin{equation}
\mL_{\gH} = \begin{bmatrix}
\diag(\mH\textbf{1}_{m}) & -\mH \\
-\mH^T & \diag(\mH^{T}\textbf{1}_{n})
\end{bmatrix},
\label{eq:laplacian}
\end{equation}
where $\mL_{\gH}\in\sR^{(n+m)\times(n+m)}$, $\diag(\cdot)$ is a mapping that converts a vector into a diagonal matrix, and $\textbf{1}_{n}\in\{1\}^{n}$ and $\textbf{1}_{m}\in\{1\}^{m}$ are two all-one vectors. Furthermore, the features of nodes in $\gV$ are $\mX_{\gV}$, and the features of nodes in $\V'$ are $\mX_{\gE}$ (as each node corresponds to a hyperedge, we use subscript $\gE$ for notational convenience).~\Cref{fig:pf_pipeline} shows an example of the incidence graph.

\begin{figure*}[t] 
\centering
\includegraphics[width=\textwidth]{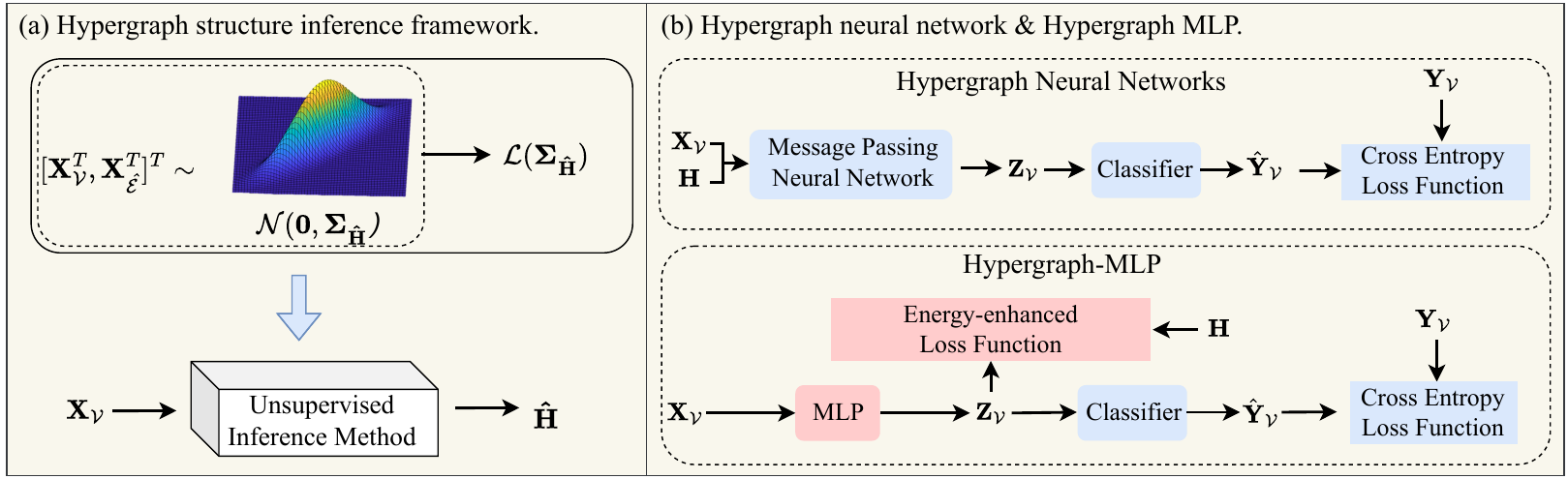}
\caption{(a) The proposed unsupervised inference method, which infers the incidence matrix of a potential hypergraph structure $\mH^*$ from given node features $\mX_{\V}$. 
(b) The training paradigms of the Hypergraph-MLP and existing hypergraph neural networks for hypergraph node classification, where $\mX_{\V}$ denotes node features, $\mH$ is a pre-defined incidence matrix, $\mY_{\gV}$ denotes node labels and $\hat{\mY}_{\gV}$ is the model predicted labels.}
\label{fig:use_cases}
\end{figure*}

\textbf{Probabilistic formulation.} We denote the conditional distribution of node and hyperdge features $\mX_{\gH} = [\mX_{\V}^{T}, \mX_{\E}^{T}]^{T}\in\sR^{(n+m)\times d}$ given incidence graph $\mH$ as $P(\mX_\gH \mid \mH)$. Following the energy-based formulation of Markov random fields, the probability can be written as,
\begin{equation}
P(\mX_\gH \mid \mH) \propto \exp \left(-f(\mH, \mX_\gH)\right).
\end{equation}
where $f(\mH, \mX_\gH)$ is the energy function. Notably, the energy function in this model can be defined as a quadratic form involving $\mX_{\gH}$ and $\mL_\gH$: 
\begin{equation}
\label{eq: energy_function}
\begin{aligned}
f(\mH, \mX_\gH) & = \operatorname{Tr}(\mX_{\gH}^\top \mL_\gH \mX_\gH) \\
&= \sum_{i=1}^{m}\sum_{v_j\in e_i}w_i~||\vx_{e_i}-\vx_{v_j}||^{2}_2
\end{aligned}
\end{equation}
where $\operatorname{Tr}$ is the trace operator. It is noted that a lower value of the energy function corresponds to a higher probability. The definition of the energy function follows our assumption introduced earlier, which gives high probability when the features of the nodes linked to a hyperedge in the incidence graph are close, and also similar to the corresponding hyperedge features. 

Following the graph machine learning literature for vanilla graphs~\cite{kalofolias2017large,dong2016learning,chepuri2017learning,pu2021learning}, we assume each dimension of $\mX_\gH$, $[\mX_\gH^\top]_{k}\in \mathbb{R}^{n}$, to be independent and identically distributed, which leads to the following decomposition,
\begin{equation}
\begin{aligned}
 P([\mX_\gH \mid \mH)&= \Pi _{k\in[1:d]}P([\mX_\gH^\top]_{k} \mid \mH),
\\
\text{with }P([\mX_\gH^\top]_{k} \mid \mH) &\propto \exp \left(-[\mX_\gH^\top]_{k}\mL_\gH[\mX_\gH^\top]_{k}^\top)\right).
\end{aligned}
\end{equation}
It can be shown that the distribution of features can be modelled by a multivariate Gaussian distribution whose covariance matrix is the pseudoinverse of the corresponding incidence graph Laplacian matrix. Therefore the conditional distribution of $P([\mX_\gH^\top]_{k} \mid \mH)$ can be modelled as:
\begin{equation}
[\mX_\gH^\top]_{k} = [[\mX_\gV^\top]_{k}^\top, [\mX_\gE^\top]_{k}]\sim \N(\textbf{0}, \mL_{\gH}^{\dagger}),
\label{eq:gaussian}
\end{equation}
where $\mL_{\gH}^{\dagger}$ is the pseudoinverse of $\mL_{\gH}$. As $\mL_{\gH}$ is uniquely associated with $\mH$, the relationship between $\mX_{\V}$ and $\mH$ is captured by~\Cref{eq:gaussian}. We name the model introduced in~\Cref{eq:gaussian} as the HMRF, which is analogous to the concept of Markov random field used to model the graph generation process, as they share the same analytical form~\cite{doi:10.1137/21M1395351, zhu2003semi}.

\subsection{An estimate for the energy function}
\label{subsec:snllf}
The probabilistic formulation in Eq.~(\ref{eq:gaussian}) relies on hyperedge features missing $\mX_{\E}$, which may not be explicitly present or observable in practical applications~\cite{chien2022you, feng2019hypergraph}. Consequently, to improve the applicability of the proposed hypergraph Markov random field, we develop an energy estimate that serves as a lower bound of the energy function but can be computed without hyperedge features. The estimate helps us approximate the likelihood function of the hypergraph structure in the structure inference task and the prior function of latent node embeddings in the node classification task.

To derive the estimate of the energy function and remove the dependency on hyperedge features, we rewrite~\Cref{eq: energy_function} as:
\begin{equation}
\label{eq:fwv}
f(\mH, \mX_{\gH}) = \sum_{i=1}^{m}\sum_{v_j\in e_i}w_i~||\vx_{e_i}-\vx_{v_j}||^{2}_2=\vw^T\vs_{\mH},
\end{equation}
where $\vw\!=\![w_1, w_2, \cdots, w_{m}]^{T}\!\in\![0,1]^{m}$ is the vectorised weights, $\vs_{\mH}\!=\![s_{e_1}, s_{e_2},\cdots,s_{e_{m}}]^T\!\in\!\sR^{m}$, and $s_{e_i}\!=\!\sum_{v_j\in e_i}||\vx_{e_i}-\vx_{v_j}||^{2}_2$. To solve the challenge of directly using $\mX_{\gV}$ to estimate $P(\mX_\gH \mid \mH)$ when $\mX_{\E}$ is missing, we further design our energy estimate as:
\begin{equation}
\label{eq: energy_estimate}
g(\mH, \mX_{\V}) = \sum_{e_i\in\E}w_i\mathop{\max}\limits_{v_j, v_k \in e_i}(||\vx_{v_j} - \vx_{v_k}||_{2}) = \vw^{T}\vs^{\prime}_{\mH},
\end{equation}
where $\vs^{\prime}_{\mH} = [s^{\prime}_{e_1}, s^{\prime}_{e_2},\cdots,s^{\prime}_{e_{m}}]^T\in\sR^{m}$, and $s^{\prime}_{e_i}=\mathop{\max}\limits_{v_j, v_k \in e_i}(||\vx_{v_j} - \vx_{v_k}||_{2}^{2})$ is the largest squared $\ell_2$ distance between any two nodes in $e_i$. We prove that $g(\mH, \mX_{\V})$ is a lower bound for $f(\mH, \mX_{\gH})$ as follows:
\begin{theorem}
\label{the:lb}
For any hypergraph structure $\mH\in\{0,1\}^{n\times m}$, given node features $\mX_\gV = [\vx_{v_1}, \vx_{v_2}, \cdots, \vx_{v_n}]^{T} \in \sR^{n\times d}$, and hyperedge features $\mX_{\E} = [\vx_{e_1}, \vx_{e_2}, \cdots, \vx_{e_{m}}]^{T} \in \sR^{m\times d}$, $g(\mH, \mX_{\V})$ is a lower bound for $f(\mH, \mX_{\gH})$.
\end{theorem}
\begin{proof}
To prove $g(\mH, \mX_{\V})$ is a lower bound for $f(\mH, \mX_{\gH})$, it suffices
to prove, for any hyperedge $e_i$ in $\mH$, that the following inequality holds:
\begin{equation}
\setlength{\abovedisplayskip}{4pt}
\setlength{\belowdisplayskip}{4pt}
\label{eq:ineq}
 \sum_{v_j\in e_i}||\vx_{e_i}-\vx_{v_j}||_2  \geq \mathop{\max}\limits_{v_j, v_k \in e_i}(||\vx_{v_j} - \vx_{v_k}||_{2}).
\end{equation} 
Without loss of generality, in the feature space, let $v_a$ and $v_b$ be the two most distant nodes within $e_i$. By triangle inequality, we have:
\begin{equation*}
\begin{aligned}
\setlength{\abovedisplayskip}{4pt}
\setlength{\belowdisplayskip}{4pt}
    \sum_{v_j\in e_i}||\vx_{e_i}-\vx_{v_j}||_2  &\geq ||\vx_{e_i}-\vx_{v_a}||_2 + ||\vx_{e_i}-\vx_{v_b}||_2 \\&\geq ||\vx_{v_a} - \vx_{v_b}||_{2}.
\end{aligned}
\end{equation*}
Hence,~\Cref{eq:ineq} holds, which proves the theorem.
\end{proof}

With the theorem, we have a theoretical guarantee to use $g(\mH, \mX_{\V})$ as a replacement for $f(\mH, \mX_{\gH})$ to solve hypergraph tasks. In~\Cref{sec:hsi}, we will show how we solve the hypergraph structure inference task through incorporating our energy estimate into~\Cref{eq: hgsi_formulation}. In~\Cref{sec:hnc}, we demonstrate that such an energy estimate can inform the design of the structure prior terms in~\Cref{eq:node_class_obj}, which lead to an efficient algorithm for the hypergraph node classification task.

\section{Hypergraph Structure Inference}
\label{sec:hsi}
In this section, we start by introducing the basic assumptions of the hypergraph structure inference task. Then, we leverage the hypergraph Markov random field introduced in~\Cref{subsec:snllf} to design a novel hypergraph structure inference framework entitled HGSI. See~\Cref{fig:use_cases} (panel a) for the illustration of this hypergraph structure inference framework.

\subsection{Assumptions}

While inferring the incidence matrix $\mH\in\{0,1\}^{n\times m}$ from observed node features $\mX_\gV$, we assume $\mH$ is drawn from a distribution that can be characterised by a weighted incidence matrix $\mH=[w_1\vh_{1}, w_2\vh_{2}, \cdots, w_{m_{p}}\vh_{m_{p}}]\in\sR^{n\times m_{p}}$, where $\vh_i\in\{0,1\}^{n}$ is one of $m_{p} = 2^n$ potential hyperedges (all possible combinations of $n$ nodes), and $w_i\in[0,1]$ defines the weight of $\vh_i$ in $\mH$. As $\vh_i$ is pre-defined, we aim to infer an optimised vector $\vw^*=[w^*_1, w^*_2, \cdots, w^*_{m_p}]^{T}\in[0,1]^{m_p}$ that contains probabilities of all potential hyperedges taking node features $\mX_\V$ as inputs, and we do so without any supservision on known ground-truth hyperedges. After $\vw^*$ is inferred, we construct $\mH^*$ by $m$ potential hyperedges with the largest weights. 


\subsection{Algorithm: Unsupervised Hypergraph Structure Inference.}
\label{subsec:learn_f}

The optimisation objective of hypergraph structure inference in~\Cref{eq: hgsi_formulation} contains the term maximising the structure likelihood $P(\mX_{\gV} \mid \mH)$ and a prior $P(\mH)$. The negative log-likelihood minimization part can be reparameterised as optimising over the energy estimate as follows,
\begin{equation}
\label{eq: energy_minimiser}
\begin{aligned}
    \argmax \log P(\mX_\gV \mid \mH) &= \argmin f(\mH, \mX_\gH) \\&\rightarrow \argmin g(\mH, \mX_\gV).
\end{aligned}
\end{equation}

With the assumption that the potential hyperedge set is fixed (the simplest case can be considering all the $2^n$ potential hyperedges), inferring $\mH^*$ from $\mX_{\V}$ is equivalent to inferring the optimal $\vw^*$ from $\mX_{\V}$. Substituting~\Cref{eq: energy_minimiser} into~\Cref{eq: hgsi_formulation} and plugging in the prior term, we can formulate the inference as the following optimisation problem:
\begin{equation}
\setlength{\abovedisplayskip}{4pt}
\setlength{\belowdisplayskip}{4pt}
\mathop{\mmin}\limits_{\vw}~\vw^{T}\vs^{\prime}_{\mH^*} - 
\alpha\textbf{1}_{m_p}^{T}\log(\vw) + \beta||\vw||_1~~\st~~w_i\in[0,1].
\label{eq:loss}
\end{equation}
where $\textbf{1}_{m_p} \in \{1\}^{m_p}$ is an all-one vector, and $\alpha$ and $\beta$ are two hyperparameters. Here the first term is the energy estimate in~\Cref{eq: energy_estimate}, which promotes hypergraph structures for which the node features satisfy our HMRF with certain regularisation terms deriving from the prior. The regularisations include: 1) a log barrier (the second term in~\Cref{eq:loss}) that enforces the positivity of the learned probabilities, which can prevent $w_i$ from being zero and thereby avoid the creation of an empty hypergraph structure; and 2) an $\ell_1$ norm (the third term)  used to ensure the sparsity of the target hypergraph structure, i.e., we assume that only a small number of potential hyperedges will be enough to explain the node features.

Taking the derivative of the objective function (denoted as $\gJ$) with respect to each $w_i\in(0,1]$:
\begin{equation*}
\frac{\partial \gJ_{wv}}{\partial w_i} =s^{\prime}_{e_i} -\frac{\alpha}{w_i} + \beta,
\end{equation*}
and setting it to zero leads to:
\begin{equation}
w^\star_i = \frac{1}{\alpha (s^{\prime}_{e_i} + \beta)},
\label{eq:opti}
\end{equation}
\begin{equation*}
\frac{\partial^{2} \gJ_{wv}(w^\star_i, \mX_{\V})}{\partial^{2} w^\star_i} = \frac{\alpha}{w^{\star^2}_i} > 0.
\end{equation*}
Hence, $\vw^{\star}=[w^{\star}_1, w^{\star}_2, \cdots, w^{\star}_{m_p}]^{T}\in(0,1]^{m_p}$ is the analytical solution to~\Cref{eq:loss} under the constraints each $w_i\in[0,1]$. In practice, we use~\Cref{eq:opti} to infer the probabilities of potential hyperedges, where we set $\alpha=1$ and $\beta=1$ to guarantee that $w_i\in[0,1]$. Note that solving~\Cref{eq:loss} does not require any labelled data, so the proposed inference method is completely unsupervised. 

\textbf{Hypergraph structure construction.} Without any constraints, $m_p=2^n$ which makes solving~\Cref{eq:loss} extremely time-consuming. To address this issue, we propose to constrain the set of potential hyperedges in $\mH^*$ in a way similar to that in~\cite{kalofolias2017large}:
for a given list $\K_S = [k_1, k_2, \cdots, k_S]$ that collects $S$ desired hyperedge sizes in descending order, each potential $k_s$-hyperedge is formed by a node with its \textit{$k_s-1$ nearest neighbours} in the feature space. By doing so, we ensure that all the potential hyperedges consist of nodes with similar features. We then use this constrained set to solve~\Cref{eq:loss} based on Algorithm~\ref{eq:opti}. Finally, we construct $\mH^*$ with $\vw^\star$. Detailed steps of our approach are provided in Algorithm~\ref{al:gis}.
\SetKwInput{KwLet}{Initialisation}
\begin{algorithm}[t]
\SetAlgoLined
\caption{Hypergraph Structure Inference (\textbf{HGSI})}
\KwIn{$\mX_{\V}$, $\K_S$, $m$, and $\m_S$ (optional; a list with number of target hyperedges in each size).}
\KwOut{Binary incidence matrix $\mH^*$.}
\KwLet{An empty set for saving inferred hyperedges $\cals$.}
\If{$\m_S$ is given} 
{
\For{$i=S:1$}
{
For $k_i$ in $\K_S$, generate potential $k_i$-hyperedges, which are formed by each node with its $k_i-1$ nearest neighbours in the feature space.

Compute $\vs^{\prime}_{\mH^*}$ by $\mX_{\V}$ for the generated $k_i$-hyperedges, and use $\vs^{\prime}_{\mH^*}$ to generate $\vw^\star$ based on~\Cref{eq:opti}.

Delete the generated $k_i$-hyperedges that are subsets of hyperedges in $\cals$, and add top-$\m_i$ hyperedges in the remaining generated $k_i$-hyperedges to $\cals$ based on $\vw^\star$.
}
Form $\mH^*$ with hyperedges in $\cals$.
}
\Else
{
\For{$i=1:S$}
{
For $k_i$ in $\K_S$, introduce potential $k_i$-hyperedges to $\cals$, which are formed by nodes with their $k_i-1$ nearest neighbours in the feature space.
}
Compute $\vs^{\prime}_{\mH^*}$ by $\mX_{\V}$ for the potential hyperedges in $\cals$, and use $\vs^{\prime}_{\mH^*}$ to generate $\vw^\star$ based on~\Cref{eq:opti}.

Form $\mH^*$ with top-$m$ potential hyperedges in $\cals$ based on $\vw^\star$.
}
\label{al:gis}
\end{algorithm}

\section{Hypergraph Node Classification}
\label{sec:hnc}
In this section, we solve the hypergraph node classification task by demonstrating a solution inspired by the hypergraph Markov random field and propose a learning framework named Hypergraph-MLP.

\textbf{Prior of latent node embeddings for hypergraph node classification.} As introduced in~\Cref{eq:node_class_obj}, we aim at injecting the inductive bias when designing the embedding prior $P(\mZ_\gH \mid \mH)$. With the data-generating process introduced in~\Cref{eq:gaussian}, we are able to explicitly constrain the embedding to follow a similar constraint as we did in HGSI. Specifically, we design the conditional probability of $P(\mZ_\gH \mid \mH)$ as: 
\begin{equation}
\begin{aligned}
&P(\mZ_{\gH}|\mH) \propto \exp\big(-f(\mH, \mZ_\gH)\big),\\
&\text{ with } f(\mH, \mZ_\gH) = \operatorname{tr } (\mZ_\gH^\top \mL_\gH \mZ_\gH).
\end{aligned}
\end{equation}
As discussed before, $\mZ_{\gE}$ does not usually exist in real-world applications. Therefore, based on~\Cref{the:lb}, we follow the same idea as above and approximate the energy function $f(\mH, \mZ_\gH)$ with $g(\mH, \mZ_\gV) = \sum_{e_i\in\E}w_i\mathop{\max}\limits_{v_j, v_k \in e_i}(||\vz_{v_j} - \vz_{v_k}||_{2}) $ defined in~\Cref{eq: energy_estimate}.

\subsection{Algorithm: Hypergraph-MLP} We start by introducing the model architecture used in our Hypergraph-MLP. Then, we explain the training process of the MLP-based model to make use of the structural information from $\mH$ without requiring it at inference time. Finally, we compare Hypergraph-MLP with existing hypergraph neural networks. To make the following discussion accessible, we denote the latent node embeddings in the layer $l$ of the MLP-based model as $\Z^{(l)}_\gV = [\z_{v_1}^{(l)}, \z_{v_2}^{(l)}, \cdots, \z_{v_n}^{(l)}]^{T} \in \sR^{n\times d}$. Following previous work~\cite{chien2022you,NEURIPS2019_1efa39bc,feng2019hypergraph, bai2021hypergraph}, we initialize $\Z_{\V}^{(0)}$ with $\mX_{\V}$. 

\textbf{Model architecture.} The training objective in~\Cref{eq:node_class_obj} suggests that the model design should consist of three components: 1) a predictor (classifier) that models $P(\mY_{l}\mid \mZ_\gH)$, 2) the prior that encodes the hypergraph structure $P\left(\mZ_\gH \mid \mH\right)$ and 3) the mapping between node features and latent embedding $P \left( \mZ_\gH \mid \mX_\gV \right)$. The overall architecture is visualised in~\Cref{fig:use_cases}. Our model first utilises a multilayer perceptron (MLP) without any message passing blocks~\cite{zhang2022graphless,hu2021graph} to encode the node embedding generation from node features, which has the form,
\begin{equation}
\label{eq:mlp}
 \Z_{\V}^{(l)} = D(LN(\sigma(\Z_{\V}^{(l-1)}\ttt^{(l)}))),
\end{equation} 
where $D(\cdot)$ is the dropout function, $LN(\cdot)$ is the layer normalization function, $\sigma(\cdot)$ is the activation function, and $\ttt^{(l)}\!\in\!\sR^{d\times d}$ denotes the learnable parameters for layer $l$. We then develop a node classifier on hypergraphs that generate the node label logits, $P(\mY_{l}\mid\mZ_\gV)$, which is formulated as:
\begin{equation}
\setlength{\abovedisplayskip}{4pt}
\setlength{\belowdisplayskip}{4pt}
\label{eq:cls}
 \hat{\mY}_{\V} = softmax(\Z_{\V}^{(L)}\mW),
\end{equation} 
where $\Z_{\V}^{(L)}\in\sR^{n\times d}$ denotes latent node embeddings generated by an $L$-layer MLP, $\hat{\mY}_{\V}\in(0,1)^{n\times c}$ denotes the node label logits, $\mW\in\sR^{d\times c}$ denotes learnable parameters, and $softmax(\cdot)$ is the softmax function. Finally, the optimisation objective also incorporated the energy estimate that is defined in~\Cref{eq: energy_estimate} for modelling $P(\mZ_\gH \mid \mH)$.

\SetKwInput{KwLet}{Initialisation}
\begin{algorithm}[t]
\SetAlgoLined
\caption{Hypergraph-MLP for Node Classification on Hypergraphs}
\textbf{/*Training*/}\\
\KwIn{$\mX_{\V}$, $\mY_{\V}$, $\mH$, $\V^{\prime}$, number of layers $L$, number of training iterations $T$.}
\KwOut{Optimal model parameters $[\ttt^{(1)^\star}, \cdots, \ttt^{(L)^\star}, \mW^{\star}]$.}
\KwLet{Randomly initialise model parameters $[\ttt^{(1)}, \cdots, \ttt^{(L)}, \mW]$, and set $\Z_{\V}^{(0)}$ as $\mX_{\V}$.}
\For{$t=1:T$ }
{
\For{$l=1:L$ }
{
Update node embeddings by~\Cref{eq:mlp}.\\
}
Generate node label logits $\hat{\mY}_{\V}$ by~\Cref{eq:cls}.\\
Update $[\ttt^{(1)}, \cdots, \ttt^{(L)}, \mW]$ by minimizing~\Cref{eq:overall_loss} with the gradient descent.
}
\textbf{/*Inference*/}\\
\KwIn{$\mX_{\V}$ and $[\ttt^{(1)^\star}, \cdots, \ttt^{(L)^\star}, \mW^{\star}]$.}
\KwOut{Predicted node label logits $\hat{\mY}_{\V}$.}
\KwLet{Set $\Z_{\V}^{(0)}$ as $\mX_{\V}$.}
\For{$l=1:L$ }
{
Update node embeddings by~\Cref{eq:mlp}.\\
}
Generate label logits $\hat{\mY}_{\V}$ by~\Cref{eq:cls}.
\label{al:hypergraph_mlp}
\end{algorithm}

\textbf{Training.} Introducing a weighting coefficient $\alpha$ in~\Cref{eq: training} yields the loss function:
\begin{equation}
\setlength{\abovedisplayskip}{4pt}
\setlength{\belowdisplayskip}{4pt}
\label{eq:overall_loss}
\ell_{\text{overall}} = \ell_{\text{CE}} + \alpha\ell_{\text{energy}}.
\end{equation}
where $\alpha$ is a hyperparameter that balances the two losses, $\ell_{\text{energy}}$ is the loss for minimising the energy function, and $\ell_{\text{CE}}$ is the cross-entropy loss for the log-likelihood $\log P(\mY_{l}\mid \mX_\gV)$ \cite{mao2023cross}.
Specifically, the energy loss is based on the energy estimate $g(\mH, \mZ_{\V})$, as defined in~\Cref{eq: energy_estimate}, 
\begin{equation}
\setlength{\abovedisplayskip}{4pt}
\label{eq:sth_loss}
\ell_{\text{energy}} = \frac{1}{m}\sum_{i=1}^{m}w_{i}\mathop{\max}\limits_{v_j, v_k \in e_i}(||\z^{(L)}_{v_j} - \z^{(L)}_{v_k}||_{2}^{2}).
\end{equation}
Minimising the energy function ensures that the output embedding of an $L$-layer MLP, $\Z_{\V}^{(L)}$, follows the DGP. In the meantime, it enables the MLP to perform inference using the structural information of the hypergraph without explicitly relying on a message passing procedure. 

Furthermore, following the previous literature~\cite{chien2022you,feng2019hypergraph}, we use the cross-entropy loss to train the overall architecture for node classificationm for approximating $P(\mY_{l}\mid \mZ_\gH)$, which is defined as:
\begin{equation}
\setlength{\abovedisplayskip}{4pt}
\label{eq:ce_loss}
\ell_{\text{CE}} = -\frac{1}{|\V^{\prime}|}\sum_{v_i\in\V^{\prime}}\y_{v_i}^{T}\log(\hat{\y}_{v_i}),
\end{equation} 
where $\V^{\prime}$ is a set of training nodes with cardinality $|\V^{\prime}|$. 

In practice, our model is trained by using gradient descent to minimize the~\Cref{eq:overall_loss}. The overall framework of our proposed Hypergraph-MLP is summarized in algorithm~\ref{al:hypergraph_mlp}. 

\textbf{Comparison with existing hypergraph neural networks.} ~\Cref{fig:use_cases} (panel b) compares the architectures of the Hypergraph-MLP and existing hypergraph neural networks. Our Hypergraph-MLP has two key benefits: 1) \textit{Inference speed.} The application of the message-passing-based hypergraph neural networks to real-world scenarios faces challenges due to high inference latency~\cite{antelmi2023survey,zhang2022graphless, hu2021graph}. Let $n$ be the number of nodes, $m$ be the number of hyperedges, and $L$ be the number of layers. The computational complexity of a hypergraph neural network is $\mathcal{O}(Ln + Lm)$, as it involves feature aggregation for every node and hyperedge in each layer. By contrast, the Hypergraph-MLP performs inference solely via feed-forward propagation, as formulated in~\Cref{eq:mlp}. Consequently, its computational complexity is $\mathcal{O}(Ln)$, which is significantly lower especially when dealing with datasets with a large number of hyperedges, such as DBLP as demonstrated in~\Cref{tab:pro_data}. 2) \textit{Inference robustness.} The significant dependence on the hypergraph structure for message passing renders current hypergraph neural networks vulnerable to structural perturbations at inference time. For instance, the introduction of random hyperedges during inference can lead well-trained hypergraph neural networks to produce suboptimal results~\cite{sun2022adversarial,hu2023hyperattack}; See Section~\ref{sec:Experiments} for empirical results. In contrast, Hypergraph-MLP implicitly takes into account the hypergraph structure, thus removing its dependence on the structure during inference. In~\Cref{sec:Experiments}, we present empirical evidence to demonstrate that this property enhances the robustness of Hypergraph-MLP compared to existing hypergraph neural networks against structural perturbations at inference.

\begin{table*}[t]
\caption{Summary of the synthetic datasets in hypergraph structure inference.}
\centering
\footnotesize\resizebox{1\columnwidth}{!}{
\begin{tabular}{@{}c|c|c|c|c@{}}
\toprule
Usage &\#Hypergraphs & \#Nodes & Hyperedge-Size & Overlap-Rate\\
\hline
\multirow{4}{*}{Study Impacts of \#Nodes} & 32&50 & 8& 30\%
\\
\cline{2-5}
 & 32&100 & 8& 30\%\\
\cline{2-5}
 & 32& 150& 8& 30\%\\
\cline{2-5}
 & 32& 200& 8& 30\%\\
\hline
\hline
\multirow{4}{*}{Study Impacts of Hyperedge-Size} & 32& 100& 4& 30\%
\\
\cline{2-5}
 &32& 100& 6& 30\%\\
\cline{2-5}
 &32& 100& 8& 30\%\\
\cline{2-5}
 &32& 100& 10& 30\%\\
\hline
\hline
\multirow{4}{*}{Study Impacts of Overlap-Rate} & 32& 100& 8&10\%
\\
\cline{2-5}
 & 32& 100& 8&30\%\\
\cline{2-5}
 & 32& 100& 8&50\%\\
\cline{2-5}
 & 32& 100& 8&70\%\\
\hline
\hline
\multirow{6}{*}{Study Impacts of \#Hyperedge-Size} & 32& 100&8 &10\%
\\
\cline{2-5}
& 32& 100&7,8,9 &10\%\\
\cline{2-5}
& 32& 100&8 &30\%\\
\cline{2-5}
& 32& 100&7,8,9 &30\%\\
\cline{2-5}
& 32& 100&8 &50\%\\
\cline{2-5}
& 32& 100&7,8,9 &50\%\\
\bottomrule
\end{tabular}}
\label{tab:syn_data_sum}
\end{table*}
\section{Experiments}
\label{sec:Experiments}
\begin{table*}[t]
\caption{F1-Score of methods on synthetic datasets with different numbers of hyperedge sizes and overlapping rates.}
\footnotesize\resizebox{\columnwidth}{!}{
\begin{tabular}{@{}lcccccc@{}}
\toprule
\multirow{2}{*}{Models / Datasets} & \multicolumn{2}{c}{Overlap Rate10\%} &   \multicolumn{2}{c}{Overlap Rate 30\%} &          \multicolumn{2}{c}{Overlap Rate 50\%}\\
& UniHG & MultiHG & UniHG & MultiHG & UniHG & MultiHG \\
\midrule
GroupNet~\cite{xu2022GroupNet} & 0.8676 & 0.2166 & 0.6059 & 0.1985 & 0.5260 & 0.1935 \\ 
HGSL~\cite{tang2023learning}  & 0.9173 $\pm$ 0.0024 & 0.8144 $\pm$ 0.0000 & 0.4754 $\pm$ 0.0044 & 0.4494 $\pm$ 0.0024 & 0.2335 $\pm$ 0.0018 & 0.1883 $\pm$ 0.0028 \\ 
NEO~\cite{8425790} &  0.9063 $\pm$ 0.0482 & 0.8666 $\pm$ 0.0142 & 0.4631 $\pm$ 0.0220 & 0.4105 $\pm$ 0.0132 & 0.1578 $\pm$ 0.0155 &  0.1425 $\pm$ 0.0094\\  
HGSI &\textbf{1.0000} &\textbf{1.0000} & \textbf{0.9301}&\textbf{0.9112} & \textbf{0.9049}& \textbf{0.8984}\\ 
\bottomrule
\end{tabular}}
\label{tab:compare_syn}
\end{table*}
\begin{table*}[t!]
\caption{HGMSE of methods on synthetic datasets with different numbers of hyperedge sizes and overlapping rates.}
\footnotesize\resizebox{\columnwidth}{!}{
\begin{tabular}{@{}lcccccc@{}}
\toprule
\multirow{2}{*}{Models / Datasets} & \multicolumn{2}{c}{Overlap Rate10\%} &   \multicolumn{2}{c}{Overlap Rate 30\%} &          \multicolumn{2}{c}{Overlap Rate 50\%}\\
& UniHG & MultiHG & UniHG & MultiHG & UniHG & MultiHG \\
\midrule
GroupNet~\cite{xu2022GroupNet} & 0.1376 &  0.5711 & 0.1815 & 0.6047 & 0.3118 & 0.6218 \\ 
HGSL~\cite{tang2023learning}  & 0.0146 $\pm$ 0.0007 & 0.0306 $\pm$ 0.0000 & 0.1231 $\pm$ 0.0018 & 0.1891 $\pm$ 0.0015 & 0.7357 $\pm$ 0.0004 & 0.6341 $\pm$ 0.0007 \\ 
NEO~\cite{8425790} &0.0445 $\pm$ 0.0004& 0.0406 $\pm$ 0.0024 & 0.6504 $\pm$ 0.0872 &  0.8826 $\pm$ 0.0336 & 1.0691 $\pm$ 0.0634 & 2.1412 $\pm$ 0.0413\\ 
HGSI  & \textbf{0.0000} & \textbf{0.0000} & \textbf{0.0155} & \textbf{0.0193} & \textbf{0.0230} & \textbf{0.0277}\\ 
\bottomrule
\end{tabular}}
\label{tab:HGSME_syn}
\end{table*}

\begin{figure*}[t!]
\centering
    \begin{subfigure}[b]{0.32\textwidth}
        \includegraphics[width=\textwidth]{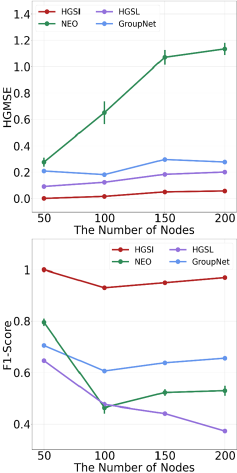}
        \caption{Impacted by the number of nodes.}
         \label{fig:num_node_fg}
    \end{subfigure}
    \hfill
    \begin{subfigure}[b]{0.32\textwidth}
        \includegraphics[width=\textwidth]{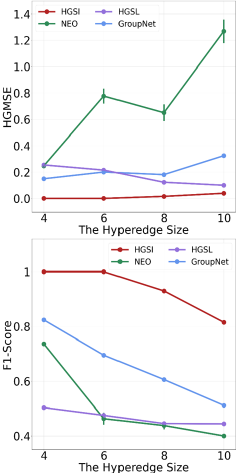}
        \caption{Impacted by the hyperedge size.}
         \label{fig:edge_size_fg}
    \end{subfigure}
    \hfill
    \begin{subfigure}[b]{0.32\textwidth}
        \includegraphics[width=\textwidth]{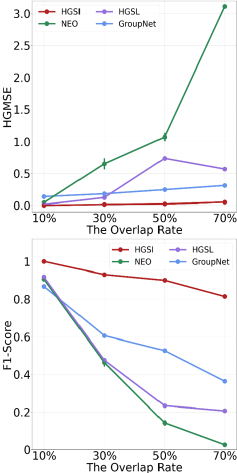}
        \caption{Impacted by the overlap rate.}
         \label{fig:over_rate_fg}
    \end{subfigure}
\caption{The impacts brought by the properties of the ground-truth hypergraph structure.}
\label{fig:syn_plot}
\end{figure*}

\begin{table*}[t]
    \begin{minipage}{\columnwidth}
      \centering
      \caption{F1-score of different methods in inferring real-world hypergraph structures.}
\label{tab:real_world_f1}
\footnotesize\resizebox{\columnwidth}{!}{
\begin{tabular}{@{}lccc@{}}
\toprule
Models / Datasets & SubCora & SubDBLP& SubYelp\\
\midrule
GroupNet~\cite{xu2022GroupNet} & 0.5310 & 0.8849 &0.8898 \\
HGSL~\cite{tang2023learning} & 0.4297 $\pm$ 0.0001 &0.4698 $\pm$ 0.0000  & 0.7215 $\pm$ 0.0010\\
NEO~\cite{8425790}& 0.4338 $\pm$ 0.0138 &0.4474 $\pm$ 0.0103  &0.7283 $\pm$ 0.0085 \\
HGSI& \textbf{0.8909}& \textbf{0.9151}& \textbf{0.9250}\\
\bottomrule
\end{tabular}}
    \end{minipage}
    \begin{minipage}{\columnwidth}
      \centering
\caption{HGMSE of different methods in inferring real-world hypergraph structures.}
\label{tab:real_world_gmse}\footnotesize\resizebox{\columnwidth}{!}{
\begin{tabular}{@{}lccc@{}}
\toprule
Models / Datasets & SubCora & SubDBLP& SubYelp\\
\midrule
GroupNet~\cite{xu2022GroupNet} & 0.4015 & 0.0758 &0.0771 \\
HGSL~\cite{tang2023learning}& 0.2001 $\pm$ 0.0001&0.0895 $\pm$ 0.0000& 0.1401$\pm$ 0.0000\\
NEO~\cite{8425790}&0.6069 $\pm$ 0.0206&0.4431 $\pm$ 0.0164&0.0915 $\pm$ 0.0140\\
HGSI & \textbf{0.0947}& \textbf{0.0425}& \textbf{0.0333}\\
\bottomrule
\end{tabular}}
    \end{minipage}
\end{table*}

\begin{figure*}[t]
\centering
\hfill
\begin{minipage}{0.4\textwidth}
\includegraphics[width=.9\textwidth,height=5.5cm]{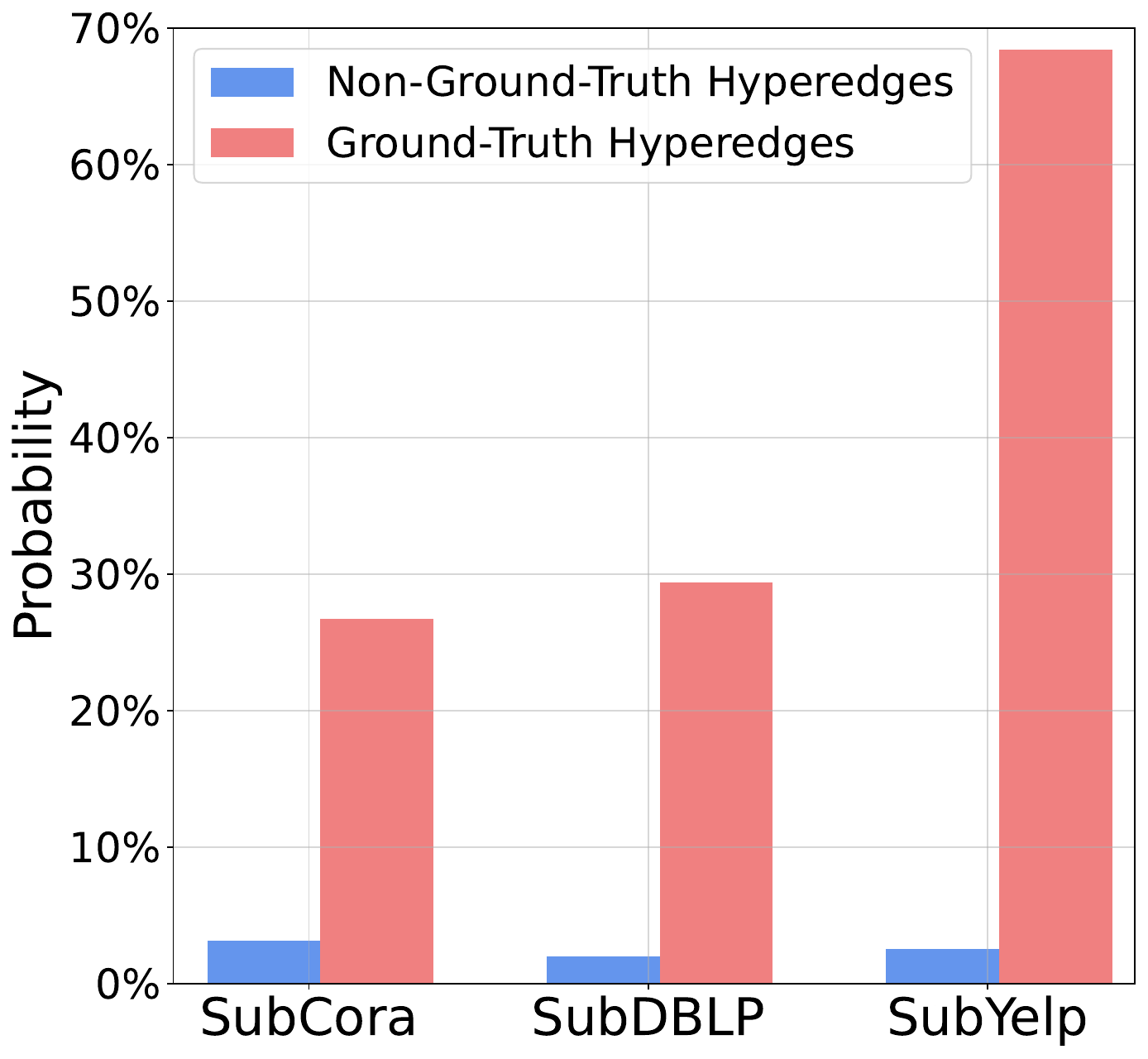}
\caption{The red/blue bar is the average probabilities generated by HGSI for the ground-truth/non-ground-truth potential hyperedges
.}  
\label{fig:pro}
\end{minipage} 
\hfill
\begin{minipage}{0.23\textwidth}
\includegraphics[width=.9\textwidth,height=5.5cm]{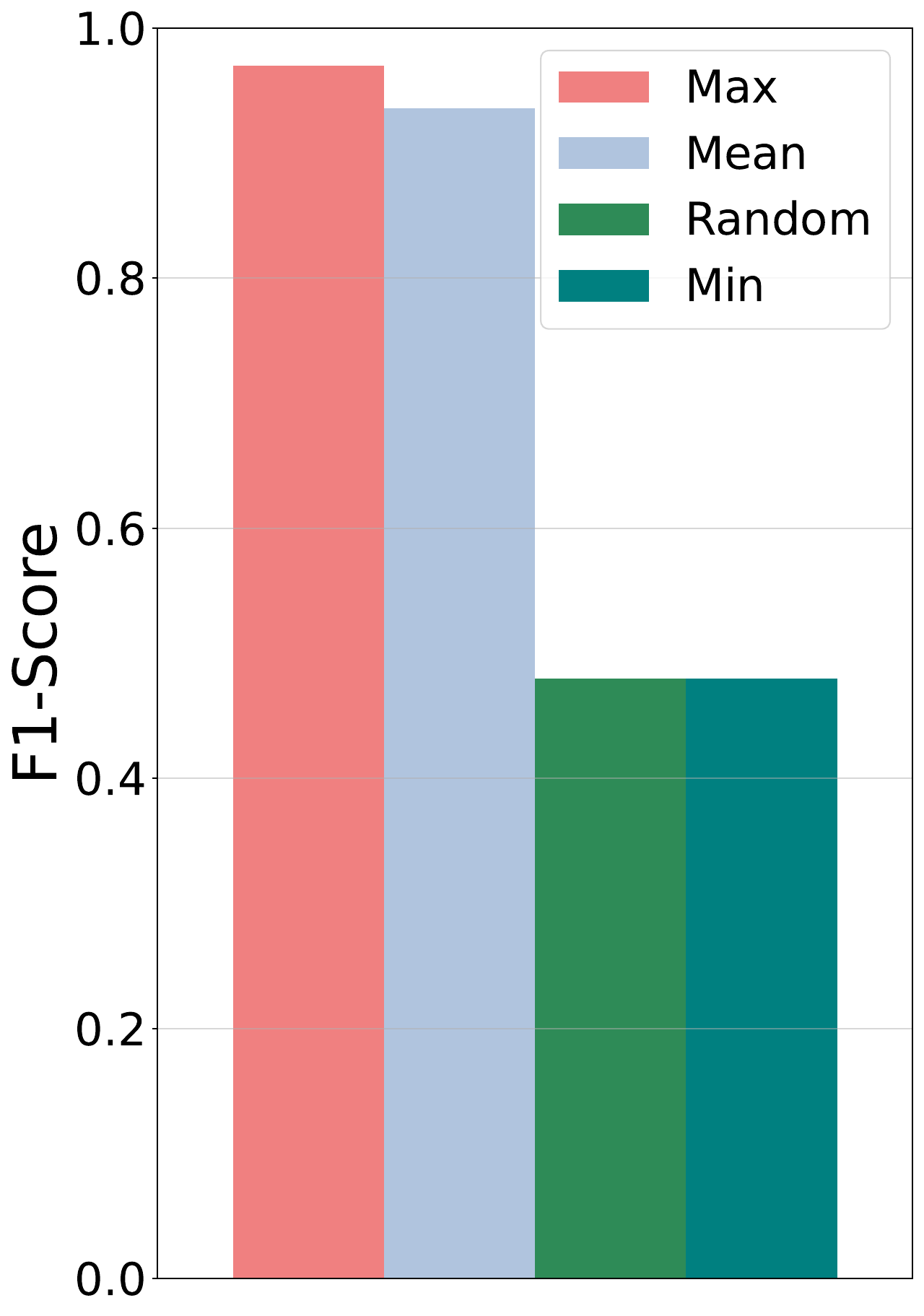}
\caption{Compare different designs of energy estimate on synthetic data.}
\label{fig:syn_sth}
\end{minipage} \hfill
\begin{minipage}{0.33\textwidth}
\includegraphics[width=\textwidth,height=5.8cm]{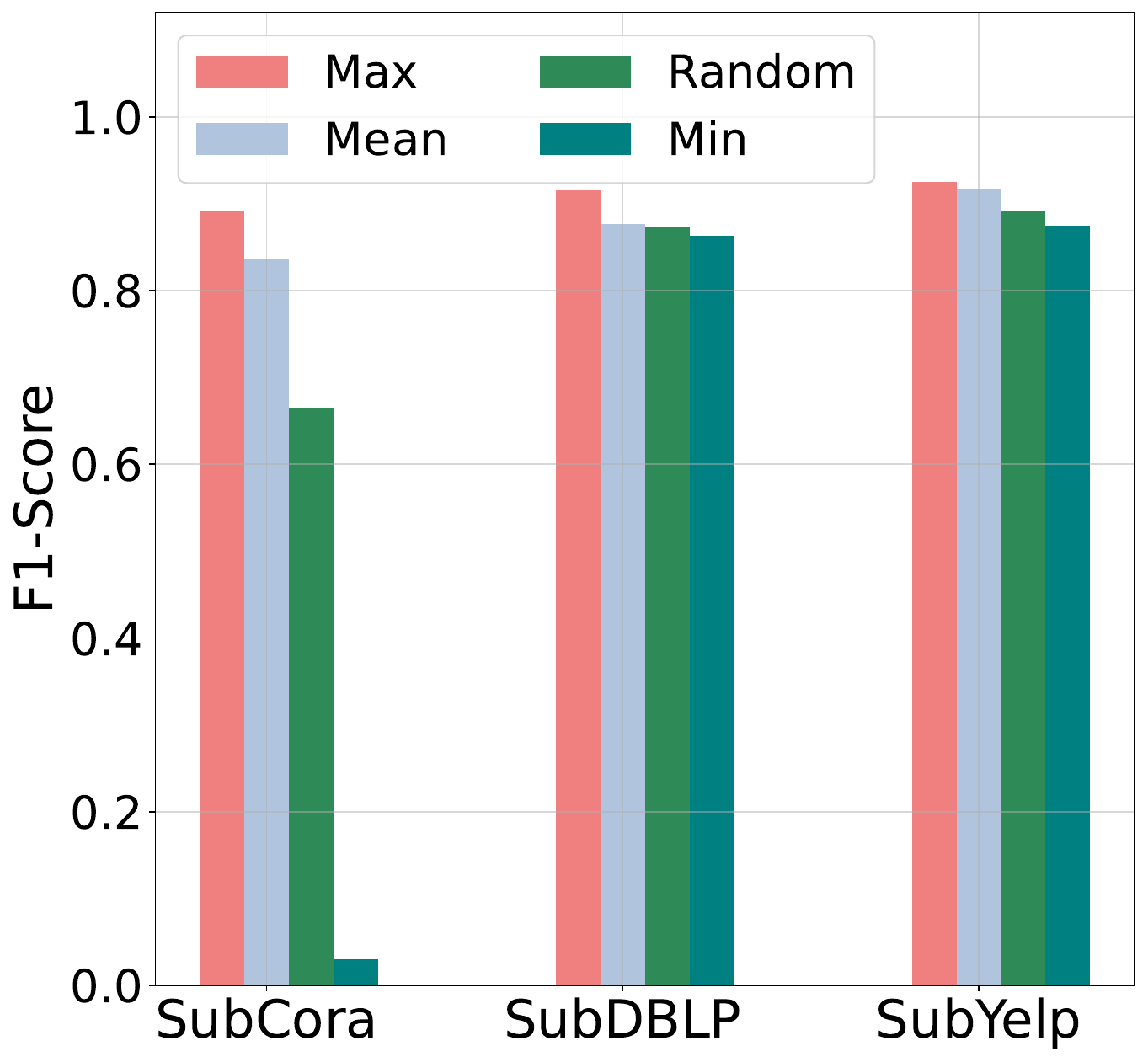}
\caption{Compare different designs of energy estimate on real-world data.}   
\label{fig:real_sth}
\end{minipage} 
\end{figure*}
In this section, we empirically evaluate our HMRF-based approaches, namely HGSI and Hypergraph-MLP. We assess HGSI on the hypergraph structure inference task and Hypergraph-MLP on the hypergraph node classification task.

\subsection{Hypergraph Structure Inference}
\textbf{Datasets.} In the \textit{synthetic dataset}, a hypergraph with node features is generated in two steps: 1) generate the ground-truth hypergraph structures based on the algorithm in~\cite{do2020structural}; 2) use the generated structures to get node features by $[\mX_{\V}^{T}, \mX_{\E}^{T}]^{T}\sim\N(\textbf{0}, (\mL_{\gH} + \sigma^{2}\textbf{I})^{-1})$
, where we set the dimension of the node features as 1000, and $\sigma$ is a small positive constant which is set to $10^{-3}$ in our experiments. 
For the synthetic data, we assume that we know the target number of hyperedges belonging to different sizes. We also define the overlap rate of a hyperedge as the ratio of nodes in the hyperedge that are involved in more than one hyperedge to the total number of nodes. The average overlapping rate of a hypergraph is the mean of its hyperedges' overlapping rates~\cite{tang2023learning}. For instance, the overlap rate of the hypergraph structure shown in Figure~\ref{fig:pf_pipeline} is 33.3\%. Here, we let all hyperedges in a hypergraph be 8-hyperedges and fix the overlap rate of each hypergraph at 30\%. The datasets used in our synthetic experiments are summarized in Table~\ref{tab:syn_data_sum}.

The \textit{real-world datasets} involve three real-world hypergraphs: SubCora, SubDBLP, and SubYelp, which are obtained from the pre-processed data in~\cite{chien2022you}. SubCora and SubDBLP are two co-authorship hypergraphs, where nodes represent authors, hyperedges represent co-authored papers, and node features are formulated by the bag-of-words model with keywords related to the research interests of each author. SubYelp is a customer hypergraph, where nodes represent customers, hyperedges representing customers of restaurants, and node features are formulated by the bag-of-words model with keywords describing the dining preferences of each customer. SubCora contains 479 nodes each with a feature dimension of 1433 and 220 hyperedges whose sizes are between 3 and 8. SubDBLP has 626 nodes each with a feature dimension of 1425 and 212 hyperedges whose sizes are 4. SubYelp includes 688 nodes each with a feature dimension of 1860 and 120 hyperedges whose sizes are 6. 
For the real-world data, we assume that we only know the target number of hyperedges. 

\textbf{Baselines.} We choose GroupNet~\cite{xu2022GroupNet}, HGSL~\cite{tang2023learning}, and NEO~\cite{8425790} as our baselines, which are all unsupervised approaches. GroupNet assumes that each node contributes to at least one hyperedge, and features of nodes in the same hyperedge are highly correlated in terms of cosine similarity. HGSL learns the hypergraph structure from node features via community detection on a learnable line graph. NEO is an overlapping community detection method based on the the $k$-means algorithm, in which generated clusters are set as hyperedges in the hypergraph.

\textbf{Metrics.} We compare the proposed approach and its variation against baselines in recovering the binary incidence matrix encoding the ground-truth hyperedges. We evaluate performance using two metrics: 1) F1‑score: Treating each potential incidence entry as a binary classification (1 = “edge present”, 0 = “edge absent”), we compute precision and recall over all node‑hyperedge pairs, then take their harmonic mean as F1‑score. A higher F1‑score indicates more accurate recovery of the incidence matrix, i.e. better classification of true edge memberships; and 2) Normalised mean squared error for hypergraph recovery (HGMSE): Measures how close the predicted incidence matrix values are to the ground truth in an MSE sense, normalised appropriately. Lower HGMSE indicates more accurate reconstruction.
In the following, we show the results of the average F1-Score/HGMSE and the associated standard deviation in ten runs. Note that the performances of NEO and HGSL vary according to different random seeds, because NEO needs to randomly initialise the cluster centres and HGSL needs to carry out random selection in its community detection step. The inference processes of GroupNet and HGSI are deterministic, so their performances do not vary according to random seeds.

\textbf{Implementation details.}  All experiments were performed on a Linux server with 96 Intel(R) Xeon(R) Gold 6248R CPU @ 3.00GHz using PyTorch. 

\textbf{Comparison with baselines on synthetic data.} We use the synthetic data to demonstrate that HGSI can learn ground-truth hypergraph structures with varying properties from data that are generated in consistency with the proposed HMRF model. We focus on four key structural properties: the number of nodes, the hyperedge size, the overlap rate, and the number of hyperedge sizes. Firstly, Fig.~\ref{fig:num_node_fg} shows the impact of the number of nodes in the ground-truth hypergraphs. 
The increase in the number of nodes would cause an increase in the search space of the tested methods. According to the structure construction constraint proposed in Section~\ref{subsec:learn_f}, the search space of our methods grows linearly with the number of nodes. Therefore, HGSI is clearly less influenced by the increase in the number of nodes. Secondly, Fig.~\ref{fig:edge_size_fg} exhibits the impact of the size of hyperedges in the ground-truth hypergraphs. 
The growing sizes of hyperedges necessitate inference techniques that incorporate a more significant number of node features to identify relevant hyperedges, resulting in higher inference complexity. Thus, the F1-Score for all employed methods declines as the hypergraph size within the ground-truth structure expands. Thirdly, Fig.~\ref{fig:over_rate_fg} displays the impact of the overlap rate in the ground-truth hypergraphs. 
The increasing overlap rate in ground-truth hypergraphs can make nodes in different hyperedges have similar features, thus leading to a blurring of the boundaries between different hyperedges. Therefore, this exacerbates the challenge encountered during the inference process. Accordingly, the performances of all the methods drop when the overlap rate increases. Finally, Table~\ref{tab:compare_syn} and Table~\ref{tab:HGSME_syn} show the impact of the number of hyperedge sizes under different overlap rates. Here each hypergraph consists of 100 nodes. Moreover, UniHG denotes hypergraphs containing hyperedges in only size 8, while MultiHG represents hypergraphs including hyperedges in three different sizes: 7, 8, and 9. We note that this metric mainly influences the performance of GroupNet. This is because the number of hyperedges in its output increases significantly when the number of hyperedge sizes grows, which results in a substantial decrease in its performance. 

\textbf{Comparison with baselines on real-world data.} The performance of each method in inferring the real-world hypergraph structures from node features is presented in Table~\ref{tab:real_world_f1} and Table~\ref{tab:real_world_gmse}. These results show that the proposed HGSI achieve state-of-the-art performance in inferring every real-world hypergraph structure with respect to both F1-Score and HGMSE. Furthermore, Figure~\ref{fig:pro} visualises the probabilities generated by HGSI for the ground-truth and non-ground-truth hyperedges in the potential hypergraph structure. These illustrate that, for every real-world hypergraph, HGSI generates much higher probabilities for ground-truth hyperedges compared to non-ground-truth hyperedges in the potential hypergraph structure. Notably, on the SubYelp dataset, the true hyperedges receive much larger probabilities than on the other benchmarks. This suggests that Yelp’s co-review patterns adhere particularly closely to the structural assumptions encoded by our HMRF. These results reflect that the proposed inference method can utilise the given node features to generate reliable probabilities for potential hyperedges in practical scenarios.

\begin{table*}[t]
\begin{center}
\caption{Properties of datasets.}
\label{tab:pro_data}
\resizebox{\textwidth}{!}{
\begin{tabular}{lcccccccc}
\hline
Name& \# Nodes & \# Hyperedges & \# Features & \# Classes & Node Definition & Hyperedge Definition\\
\hline
Cora-CC  &  2708   &  1579   &    1433     &    7  & Paper & Co-citation     \\
Citeseer &  3312   &  1079   &    3703     &    6   & Paper & Co-citation    \\
Pubmed   & 19717   &  7963   &     500     &    3  & Paper & Co-citation     \\
DBLP-CA  & 41302   & 22363   &    1425     &    6 & Paper & Co-authorship      \\
20News   & 16242   &   100   &     100     &    4  & Document &  Containing the Same Word   \\
NTU2012  &  2012   &  2012   &     100     &   67  & 3D object  &  K-Nearest-Neighbors in Feature Space   \\
\hline
\end{tabular}}
\end{center}
\end{table*}

\begin{table*}[t]
\begin{center}
\caption{Comparison with baselines on clean datasets in terms of the test ACC (\%).}
\label{tab:real-world_a}\resizebox{\columnwidth}{!}{
\begin{tabular}{c|cccccc}
\hline
&Cora-CC & Citeseer & Pubmed & DBLP-CA & 20News & NTU2012
\\
\hline
HyperGCN & 78.45 ± 1.26 &
  71.28 ± 0.82 &
  82.84 ± 8.67 &
  89.38 ± 0.25 &
  81.05 ± 0.59 &
  56.36 ± 4.86 
\\
HGNN & 79.39 ± 1.36 &
  72.45 ± 1.16 &
  86.44 ± 0.44 &
  91.03 ± 0.20 &
  80.33 ± 0.42 &
  87.72 ± 1.35 
\\
HCHA & 79.14 ± 1.02 &
  72.42 ± 1.42 &
  86.41 ± 0.36 &
  90.92 ± 0.22 &
  80.33 ± 0.80 &
  87.48 ± 1.87
\\
UniGCNII & 78.81 ± 1.05 &
73.05 ± 2.21 &
  88.25 ± 0.40 &
  \textbf{91.69 ± 0.19} &
  81.12 ± 0.67 &
  \textbf{89.30 ± 1.33}
\\
AllDeepSets & 76.88 ± 1.80 &
  70.83 ± 1.63 &
  \textbf{88.75 ± 0.33} &
  91.27 ± 0.27 &
  81.06 ± 0.54 &
  88.09 ± 1.52
\\
AllSetTransformer & 78.59 ± 1.47 &
  73.08 ± 1.20 &
  88.72 ± 0.37 &
  91.53 ± 0.23 & 
  81.38 ± 0.58 &
  88.69 ± 1.24
\\
MLP & 74.99 ± 1.49&  72.31 ± 1.28&   87.69 ± 0.59& 85.53 ± 0.27&  81.70 ± 0.49&  87.89 ± 1.36
\\
\hline
Hypergraph-MLP &\textbf{79.80 ± 1.82} &  \textbf{73.90 ± 1.57}& 87.89 ± 0.55& 90.29 ± 0.26&  \textbf{81.75 ± 0.41}&  88.42 ± 1.32
\\
\hline
\end{tabular}}
\end{center}
\end{table*}

\begin{table*}[t]
\begin{center}
\caption{Comparison with baselines on clean datasets in terms of the inference time (ms).}
\label{tab:real-world_t}\resizebox{\columnwidth}{!}{
\begin{tabular}{c|cccccc}
\hline
&Cora-CC & Citeseer & Pubmed & DBLP-CA & 20News & NTU2012
\\
\hline
HyperGCN & 0.46 ±
0.10
  &  0.49 ±
0.10& 
0.60 ± 0.07 &
  1.19 ± 0.21
 & 
0.82 ± 0.30 &
  0.47 ± 0.10
\\
HGNN & 
1.40 ± 0.11&
1.41 ± 0.33&
3.33 ± 0.26 &
4.36 ± 0.41& 
1.39 ± 0.33&
1.26 ± 0.11
\\
HCHA & 1.33 ±
0.38 &1.37 ±
0.14&
  3.43 ±
0.17&
  8.08 ±
0.76& 
1.45 ±
0.14 &
  1.29 ±
0.15
\\
UniGCNII & 3.01 ± 0.14 &
 0.71 ± 0.10 &
 0.85 ± 0.07   &
 21.15 ± 0.64 & 
  3.79 ± 0.18
 &
  1.48 ± 0.08
  
\\
AllDeepSets & 2.02 ±
0.15 &
  2.40 ±
0.14 &
  8.90 ±
1.09 &
  21.51 ±
0.43 & 
 7.68 ±
0.57
 &
  1.96 ±
0.09 

\\
AllSetTransformer & 1.91 ±
1.34 &
  3.01 ±
1.32 &
  4.75 ±
1.33 &
  21.21 ±
1.42 & 
  4.30 ±
1.57
 &
  1.90 ±
1.32
\\
\hline
Hypergraph-MLP &\textbf{0.45 ± 0.11} &  \textbf{0.21 ±
0.07}& 
\textbf{0.19 ± 0.04}& \textbf{0.47 ± 0.17}&  \textbf{0.28 ±
0.05}&  \textbf{
0.36 ±
0.08}
\\
\hline
\end{tabular}}
\end{center}
\end{table*}

\textbf{Ablation study on the probabilistic criterion.} We empirically study how the probabilistic criterion adopted in the energy estimate, as defined in~\Cref{eq: energy_estimate}, influences HGSI. We denote the proposed probabilistic criterion as Max and compare it with three other probabilistic criteria with different bases: Mean, Random and Min. Mean calculates the probabilistic criterion based on the average squared $\ell_2$ distance between nodes in a hyperedge, Random uses the squared $\ell_2$ distance of a random node pair, and Min considers the smallest squared $\ell_2$ distance between nodes in the hyperedge. We conduct experiments on both synthetic and real-world data. For the synthetic data, we use the 200-node dataset with hyperedges of size 8 and a 30\% overlap rate. For the real-world data, we use SubCora, SubDBLP, and SubYelp. The results are summarized in~\Cref{fig:syn_sth} and~\Cref{fig:real_sth}, which show that the proposed probabilistic criterion enables HGSI to achieve its optimal performance. This likely occurs because the proposed probabilistic criterion corresponds to the tightest lower bound of the true HMRF energy criterion among the four options.

\subsection{Hypergraph Node Classification}

\textbf{Datasets.} We use six public datasets, including scientific collaborations (Cora-CC, Citeseer, Pubmed, and DBLP) adapted from~\cite{NEURIPS2019_1efa39bc}, 20News~\cite{dua2017uci}, and NTU2012~\cite{chen2003visual}. Details of the datasets used are included in Table~\ref{tab:pro_data}.

\begin{figure*}[t!]
\begin{center}
\centering
\begin{subfigure}[b]{0.4\textwidth}
        \includegraphics[width=\textwidth]{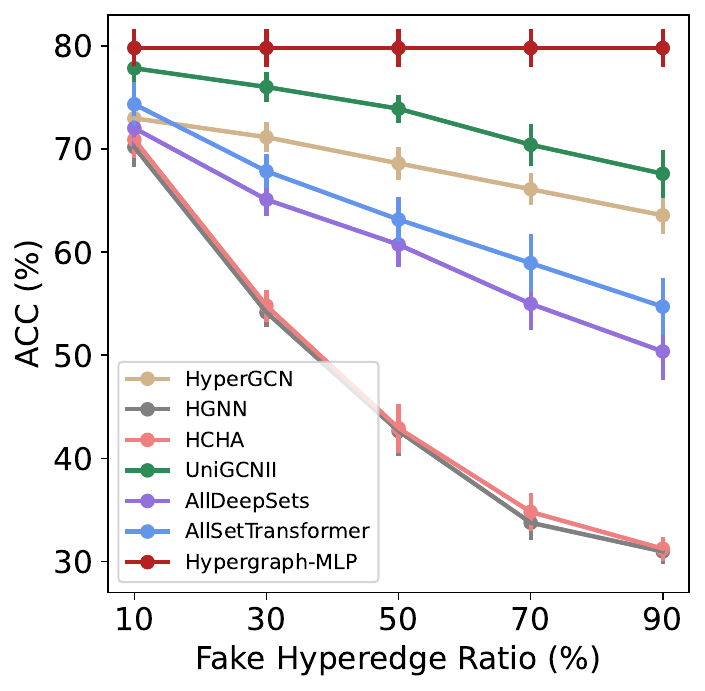}
        \caption{Cora-CC.}
\end{subfigure}
\begin{subfigure}[b]{0.4\textwidth}
        \includegraphics[width=\textwidth]{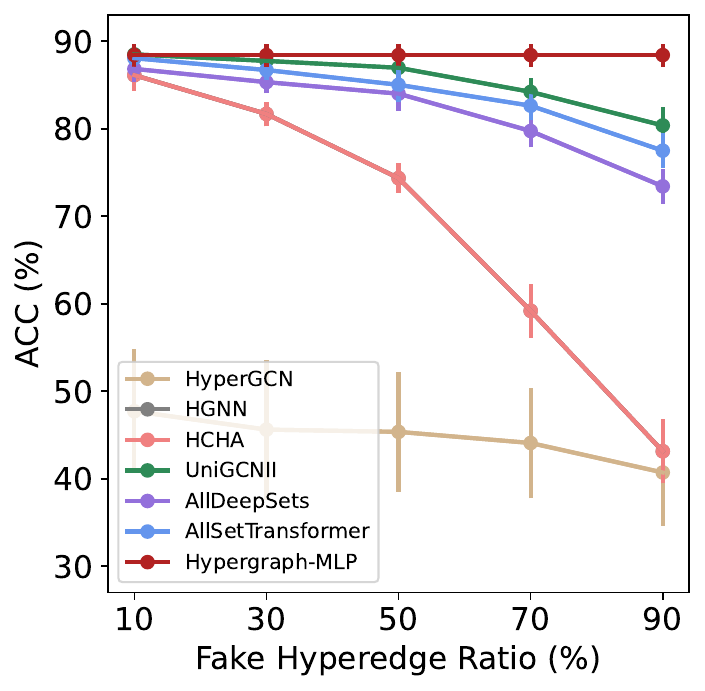}
        \caption{NTU2012.}
\end{subfigure}
\caption{Comparision with baselines on the perturbed Cora-CC and NTU2012.}
\label{fig:acc}
\end{center}
\end{figure*}

\textbf{Baselines.} We compare Hypergraph-MLP with six message-passing hypergraph neural networks: HyperGCN~\cite{NEURIPS2019_1efa39bc}, HGNN~\cite{feng2019hypergraph}, HCHA~\cite{bai2021hypergraph}, UniGCNI~\cite{ijcai21-UniGNN}, AllDeepSets~\cite{chien2022you}, and AllsetTransformer~\cite{chien2022you}; and a standard MLP, differing from our Hypergraph-MLP only in the loss function, using~\Cref{eq:ce_loss}) instead of~\Cref{eq:overall_loss}.

\textbf{Metrics.} In line with prior research~\cite{chien2022you,NEURIPS2019_1efa39bc,feng2019hypergraph,bai2021hypergraph,ijcai21-UniGNN}, we evaluate the performances of baselines and our method by accuracy (ACC: $\%$), which is defined as the ratio between the number of correct predictions to the total number of predictions.

\begin{figure*}[t!]
\begin{center}
\centering
\begin{subfigure}[b]{0.48\textwidth}
\includegraphics[width=\textwidth,height=6cm]{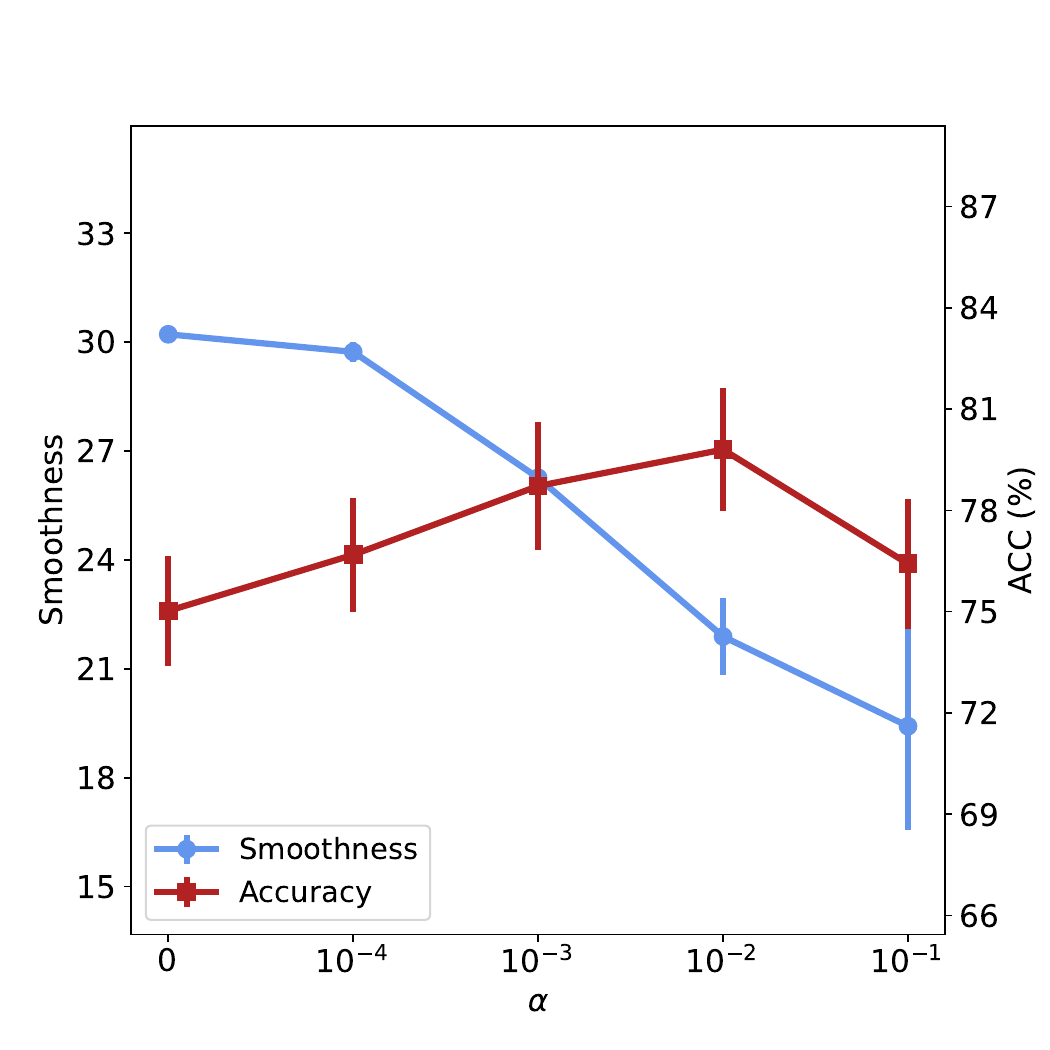}
        \caption{Cora-CC.}
\end{subfigure}
\begin{subfigure}[b]{0.48\textwidth}
\includegraphics[width=\textwidth,height=6cm]{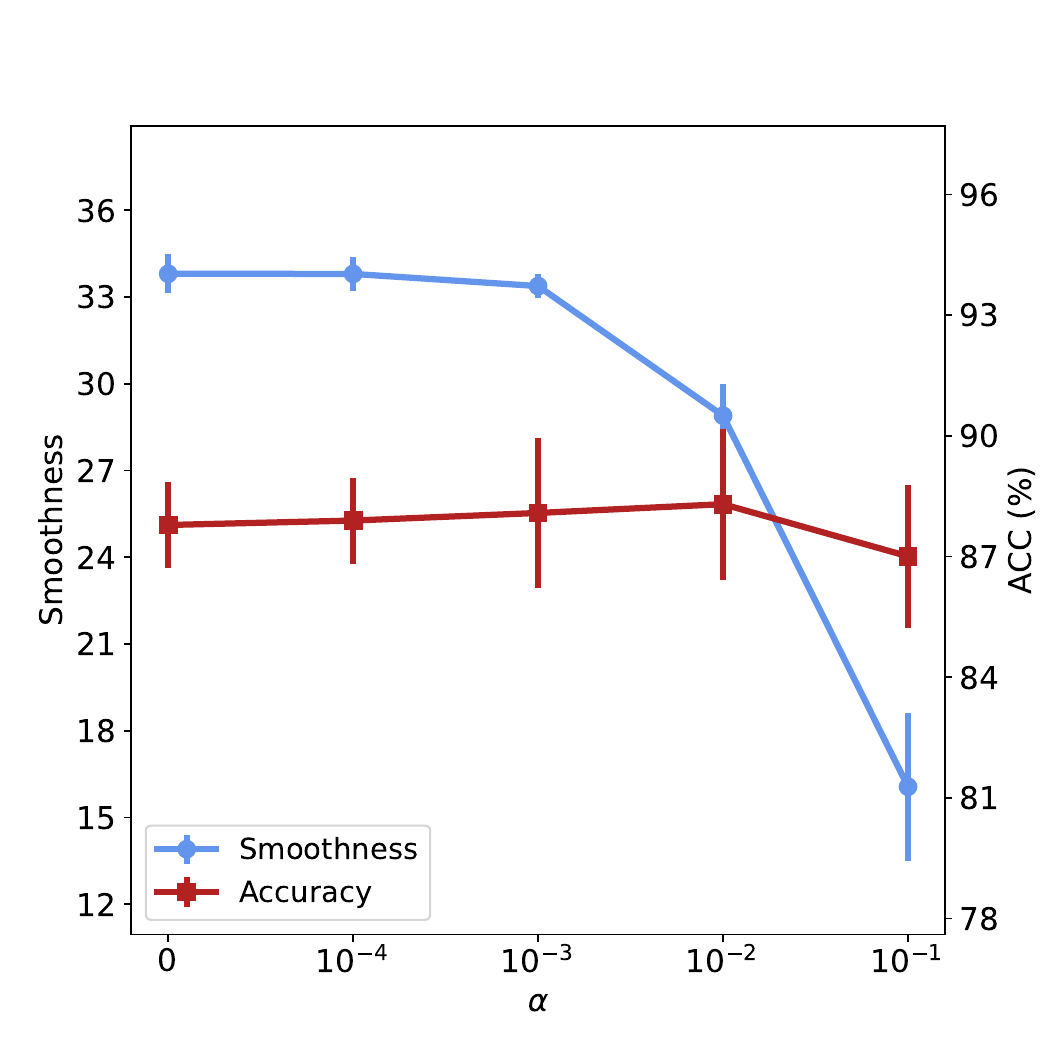}
        \caption{NTU2012.}
\end{subfigure}
\caption{Impact of hyperparameters.}
\label{fig:al}
\end{center}
\end{figure*}

\textbf{Implementation details.} Following a previous work~\cite{chien2022you}, we randomly partitioned the dataset into training, validation, and test sets using a 50\%/25\%/25\% split. All experiments were performed on an RTX 3090 using PyTorch. We identified the optimal value of $\alpha$ in~\Cref{eq:overall_loss} through grid search. 

\textbf{Comparison with baselines on clean data.} The results for real-world hypergraph node classification benchmarks, conducted without considering structural perturbations, are summarised in Table~\ref{tab:real-world_a} and Table~\ref{tab:real-world_t}. In Table~\ref{tab:real-world_a}, we compare the proposed Hypergraph-MLP against baseline methods in terms of ACC. Our results demonstrate that Hypergraph-MLP outperforms the baseline methods across three datasets (Cora-CC, Citeseer, and 20News). Furthermore, Hypergraph-MLP consistently outperforms the standard MLP across all the datasets. These findings demonstrate the efficacy of the energy-enhanced loss function in effectively enabling the MLP-based model to leverage hypergraph structural information, while not explicitly using the structure itself as input for message passing. In addition, as illustrated in Table~\ref{tab:real-world_t}, Hypergraph-MLP demonstrates the fastest inference speed on all six chosen datasets compared to existing message-passing-based hypergraph neural networks. Notably, on the Pubmed dataset, Hypergraph-MLP achieves an average mean inference time that is only $30\%$ of the time required by the fastest message-passing-based model (HyperGCN) and a mere $2\%$ of the time needed by the most time-consuming message-passing-based model (AllDeepSets). This computational benefit renders Hypergraph-MLP well-suited for applications with stringent latency constraints, in contrast to existing hypergraph neural networks.

\textbf{Comparison with baselines on perturbed data.} We summarise the results on perturbed Cora-CC and NTU2012 datasets in Fig.~\ref{fig:acc}. We perturb hypergraph structures by randomly replacing ground-truth hyperedges with random ones during inference. Moreover, we define the random hyperedge ratio as $\frac{n_{f}}{n_{o}}$, where $n_{f}$ is the number of random hyperedges in the perturbed structure and $n_{o}$ is the number of hyperedges in the ground-truth structure. The results are consistent across both datasets: compared with message-passing-based hypergraph neural networks, Hypergraph-MLP consistently demonstrates better robustness against structural perturbations. Indeed, removing the reliance on hypergraph structure during inference makes Hypergraph-MLP unaffected by structural perturbations introduced exclusively at inference time. This advantage makes Hypergraph-MLP a robust choice for applications vulnerable to structural evasion attacks.

\textbf{Impact of hyperparameters}
We finally study the impact of the hyperparameter $\alpha$ in Eq.~(\ref{eq:overall_loss}) 
Specifically, we use the Cora-CC and NTU2012 datasets to demonstrate how the energy of the generated node features and the classification accuracy of Hypergraph-MLP vary with different values of $\alpha$. In~\Cref{fig:al}, we quantify the energy estimate based on~\Cref{eq: energy_estimate}, and the result reveals that a modest $\alpha$ value facilitates Hypergraph-MLP in producing node features with a suitable level of energy, which, in turn, improve the classification accuracy.

\section{Conclusion}
\label{sec:Conclusion}
We propose a hypergraph Markov random field for hypergraph machine learning. This model encodes the data-generating processes of node and hyperedge features on a hypergraph as a multivariate Gaussian distribution, whose covariance matrix is determined by the hypergraph structure itself. Leveraging this model, we propose an energy estimate designed to approximate both the structure likelihood function and the node embedding prior in hypergraph machine learning tasks. Based on this estimate, we propose two novel frameworks for two hypergraph-based learning tasks: 
HGSI for inferring the latent hypergraph structure from node features, and Hypergraph-MLP for node classification on pre-defined hypergraphs. Extensive experiments on both tasks confirm the effectiveness and efficiency of the proposed frameworks. We believe both the proposed HMRF and the derived hypergraph machine learning frameworks can benefit the hypergraph research community from at least three distinct perspectives: 1) offering a mathematically rigorous perspective on understanding the interactions between node data and hypergraph structures; 2) providing an effective method for inferring an appropriate hypergraph structure to capture the higher-order relationships among data entities; and 3) delivering an efficient and robust learning framework for learning node embeddings with pre-defined hypergraphs.

\bibliography{mybibfile.bib}

\end{document}